\documentclass[10pt,twocolumn,letterpaper]{article}

\usepackage{iccv}
\usepackage{times}
\usepackage{epsfig}
\usepackage{graphicx}
\usepackage{amsmath}
\usepackage{amssymb}
\usepackage{pifont}

\usepackage{booktabs}
\usepackage{arydshln}
\usepackage{color}
\usepackage{colortbl}
\usepackage{xcolor} 
\usepackage{caption}
\usepackage{subcaption}
\usepackage{multirow}
\usepackage{rotfloat}
\usepackage{capt-of}
\usepackage{subcaption}
\usepackage{mwe}
\usepackage{longtable}
\usepackage{diagbox}
\usepackage{makecell}

\usepackage{algorithm}
\usepackage{algorithmic}
\usepackage{listings}

\definecolor{linkc}{rgb}{0, 0.44, 0.74}
\definecolor{eqc}{rgb}{1, 0, 0}
\usepackage[pagebackref=false,breaklinks=true,colorlinks,urlcolor=linkc,citecolor=linkc,linkcolor=eqc,bookmarks=false]{hyperref}

\def\iccvPaperID{7004}

\newcommand{\model}{MetaBEV}
\newcommand{\moe}{M\({}^{\mbox{\small2}}\)oE}
\newcommand{\cmark}{\ding{51}}
\newcommand{\xmark}{\ding{55}}

\definecolor{revise}{HTML}{0071BC}
\def\gain#1{\textbf{\textcolor{revise}{#1}}}

\iccvfinalcopy
\ificcvfinal\fi

\begin{document}

\title{MetaBEV: Solving Sensor Failures for BEV Detection and Map Segmentation}

\author{Chongjian Ge$^{1*}$ {}
  Junsong Chen$^{3,1*}$ {}
  Enze Xie$^{2\dagger}$ {}
  Zhongdao Wang$^2$  {} \\
  Lanqing Hong$^2$ {}
  Huchuan Lu$^3$  {} 
  Zhenguo Li$^2$  {}
  Ping Luo$^{1\dagger}$\\
  [0.1cm]
  $^1$The University of Hong Kong  \quad $^2$Huawei Noah's Ark Lab \quad $^3$Dalian University of Technology \\ 
  {\tt\small rhettgee@connect.hku.hk \quad jschen@mail.dlut.edu.cn \quad lhchuan@dlut.edu.cn}
 \\ {\tt\small  \{xie.enze,wangzhongdao,honglanqing,li.zhenguo\}@huawei.com \quad pluo@cs.hku.hk}
 \\ {\small Project page: \url{https://chongjiange.github.io/metabev.html}}
}

\twocolumn[{%
\maketitle
\vspace{-9mm}
\begin{figure}[H]
\hsize=\textwidth
\centering
\includegraphics[width=1.0\textwidth]{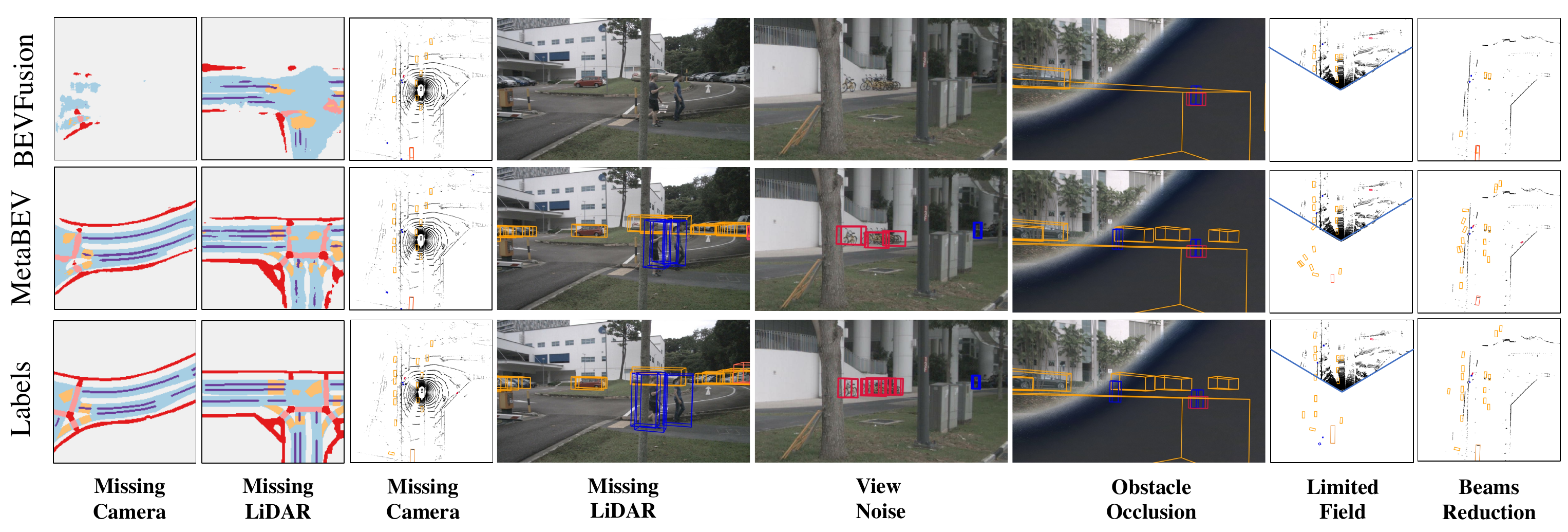}
\vspace{-5mm}
\caption{
\textbf{Selected 3D prediction results under various sensor failures.} \model{} shows stronger robustness on both map segmentation and 3D detection compared to the representative multi-modal method, \eg, BEVFusion~\cite{liu-2022-bevfusion}. Notably, in case of an entire sensor absent (e.g., Camera or LiDAR), \model{} provides satisfactory outcomes while existing methods fail to. \textbf{Top two rows}: 3D prediction results. \textbf{Bottom row}: the corresponding ground truths.
}

\label{fig:vis1}
\end{figure}
}]

\maketitle
\ificcvfinal\thispagestyle{empty}\fi

\footnotetext{$*$ Equal contribution.}
\footnotetext{$\dagger$ Corresponding authors.}

\begin{abstract}
Perception systems in modern autonomous driving vehicles typically take inputs from complementary multi-modal sensors, \eg, LiDAR and cameras. However, in real-world applications, sensor corruptions and failures lead to inferior performances, thus compromising autonomous safety. In this paper, we propose a robust framework, called MetaBEV, to address extreme real-world environments, involving overall six sensor corruptions and two extreme sensor-missing situations. 
In MetaBEV, signals from multiple sensors are first processed by modal-specific encoders. Subsequently, a set of dense BEV queries are initialized, termed meta-BEV. These queries are then processed iteratively by a BEV-Evolving decoder, which selectively aggregates deep features from either LiDAR, cameras, or both modalities. The updated BEV representations are further leveraged for multiple 3D prediction tasks. Additionally, we introduce a new \moe{} structure to alleviate the performance drop on distinct tasks in multi-task joint learning.
Finally, \model\ is evaluated on the nuScenes dataset with 3D object detection and BEV map segmentation tasks.  Experiments show \model\ outperforms prior arts by a large margin on both full and corrupted modalities.
For instance, when the LiDAR signal is missing, MetaBEV improves $35.5\%$ detection NDS and $17.7\%$ segmentation mIoU upon the vanilla BEVFusion~\cite{liu-2022-bevfusion} model; and when the camera signal is absent, MetaBEV still achieves $69.2\%$ NDS and $53.7\%$ mIoU, which is even higher than previous works that perform on full-modalities. Moreover, MetaBEV performs fairly against previous methods in both canonical perception and multi-task learning settings, refreshing state-of-the-art nuScenes BEV map segmentation with $70.4\%$ mIoU.
\end{abstract}

\section{Introduction} \label{sec:intro}
Perceiving the surrounding environment is a fundamental capability to autonomous driving systems. In pursuit of higher perceptual accuracy, prior works make significant efforts in designing stronger task-specific modules~\cite{zhou-18cvpr-voxelnet,lang-22cvpr-pointpillar,li-22eccv-bevformer}, cultivating effective training paradigms~\cite{chen-2022-co}, leveraging multi-modalities~\cite{liu-2022-bevfusion,liang-22cvpr-bevfusion,chen-2022-futr3d}, etc. Among all these, the multi-sensors fusion strategy exhibits significant advantages in achieving stronger perception abilities~\cite{vora-20cvpr-pointpainting,yin-21nips-mvp,liu-2022-bevfusion,bai-22cvpr-transfusion}, thus being widely explored in both academia and industry.
While the majority of works focus on achieving optimal performance on ideal multi-modal inputs for a single specific task, they unintentionally neglect how the designed models perform with sensor failures, which are commonly encountered and inevitable in real-world applications. 

To alleviate performance drop on sensor failures, previous works encounter two challenges as follows. 
\textbf{1) Features misalignment:} Existing fusion methods typically utilize CNNs and features concatenation for fusion~\cite{liu-2022-bevfusion,liang-22cvpr-bevfusion}. The pixel-level position correlation is consistently imposed, giving rise to multi-modal features misalignment, especially when geometric-related noises are introduced. This issue could be attributed to the intrinsic characteristics of CNNs, which exhibit limitations in long-range perception and adaptive attention to input features.
\textbf{2) Heavily reliant on complete modalities:}
Prior arts generate the fused BEV features using either query-indexing or channel-wise fusion manners. Query-indexing methods~\cite{sindagi-19icra-mvx,vora-20cvpr-pointpainting,wang-21cvpr-pointaugmenting,huang-20eccv-epnet} typically rely on LiDAR and 2D Camera features for mutual querying, while channel-wise fusion approaches~\cite{liu-2022-bevfusion,liang-22cvpr-bevfusion} are inevitable to involve element-wise operations (\eg element-sum) for feature merging. Both fusion strategies are heavily dependent on complete modality inputs and lead to inferior perception performance encountering extreme sensor failures such as LiDAR-missing or Cameras-missing, thus being limited in the practical applications.

\begin{figure}[t]
\begin{center}
\footnotesize
\includegraphics[width=1\linewidth]{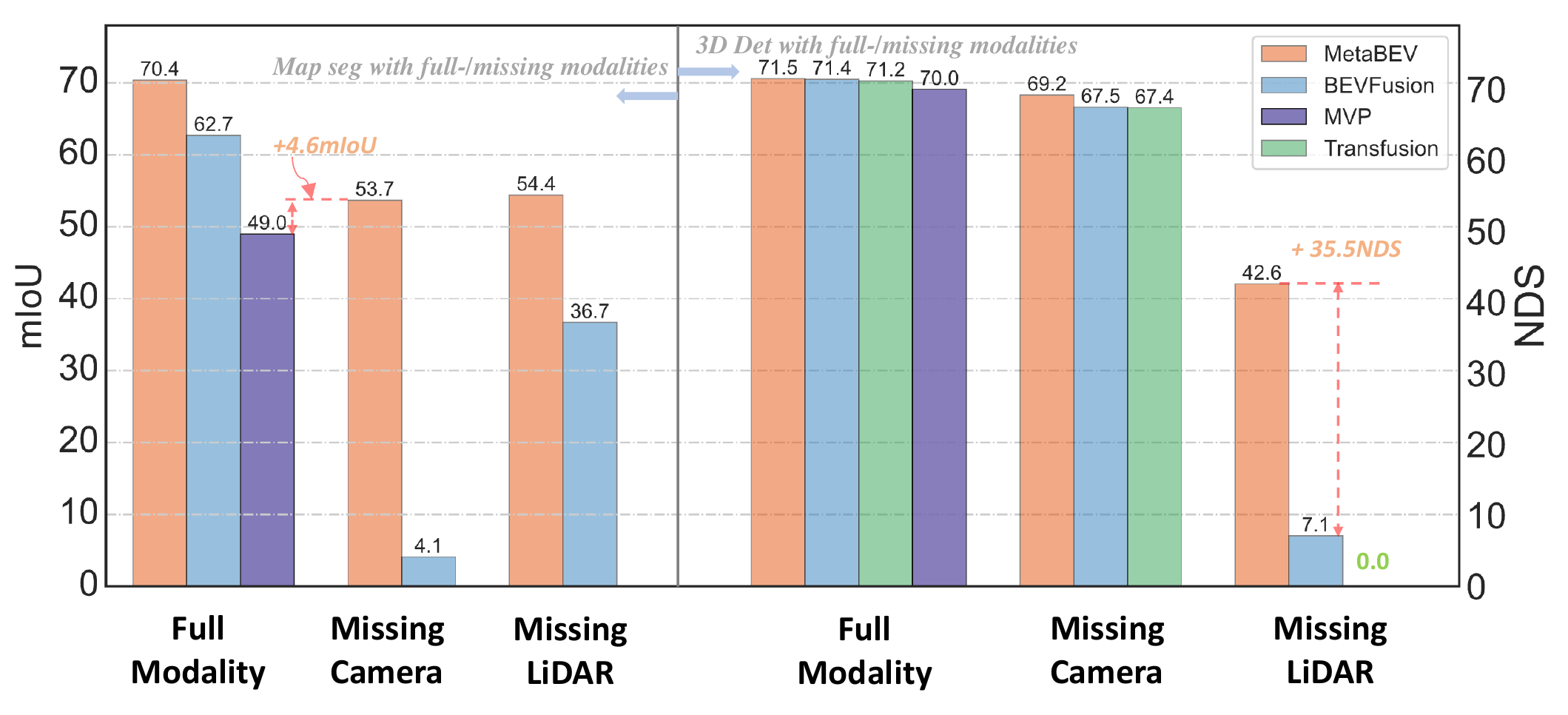}\\

\caption{\textbf{\model\ shows stronger robustness on missing sensors.} We perform representative methods with full/missing modalities on both map segmentation (the left part) and 3D detection (the right part). 
Results show \model\ can mitigate the performance drop on input absence. Quantitatively, when facing missing cameras, \model\ still achieves 53.7\% mIoU, which outperforms the representative method (i.e., BEVFusion~\cite{liu-2022-bevfusion}), by +49.6\% mIoU, and is even better than MVP~\cite{yin-21nips-mvp} performed on full-modalities. The superior performance could also be found on 3D detection tasks.
}
\vspace{-2em}
\end{center} 

\end{figure} \label{fig:teaser}

In this research, we propose \model{} to tackle the above features misalignment and full-modality dependence problems through modality-arbitrary and task-agnostic learning in the unified bird’s-eye view (BEV) representation space~\cite{li-2022-unifying}. We identify the major bottleneck in modality-dependent methods is the lack of designs that enable independent fusion of different modalities by the fusion module. Therefore, we present a modality-arbitrary BEV-Evolving decoder, which leverages cross-modal attention to correlate the learnable meta-BEV queries with either a single camera-BEV feature, LiDAR-BEV feature, or both to eliminate the bottleneck. Finally, we apply a few task-specific heads to support different 3D perception predictions.

Except for the canonical perception (on no corrupted sensors), we also evaluate \model\ on overall six sensor corruptions (Limited Field (LF), Beams Reduction (BR), Missing of Objects (MO), View Drop (VD), View Noise (VN), Obstacle Occlusion (OO)) and two sensor-missing scenarios (Missing LiDAR (ML) and Missing Camera (MC)). Compared with prior works, \model\ performs more robustly as depicted in Fig.~\ref{fig:teaser}. For example, it achieves 69.2\% NDS and 42.6\% NDS on detection when totally missing the cameras or LiDAR, respectively. 
For map segmentation, when total cameras absence occurs, \model\ could still achieve better performance compared to the work trained on multi-modalities (\ie, 54.7\%  \textit{v.s.} MVP's 49.0\% mIoU~\cite{yin-21nips-mvp}).
Besides, the attention-based \model\ exhibits inherent robustness against multiple heavy corruptions with zero-shot and in-domain tests. For instance, as shown in Tab.~\ref{tab:corruptions}, even missing 66.6\% of LiDAR points, \model\ still achieves 55\% NDS, outperforming the competitor by +11.7\%.

Moreover, considering the limited computational resource in practice, using a single framework with shared parameters for different tasks is more efficient than using separate frameworks for multiple tasks. However, the task conflicts in the joint learning of detection and segmentation often lead to severe performance drop~\cite{Xie-22arxiv-m2bev, li-22eccv-bevformer, liu-2022-bevfusion}, and existing methods rarely analyze and design for multi-task learning (MTL). We incorporate \model\ with a flexible module based on Multi-Task Mixture of Experts (\moe{}) to demonstrate one possible solution for MTL and hope to stimulate further research in this area.

The appealing advantages of \model\ are concluded:
\begin{enumerate}
\item \model\ is a novel BEV perception framework for 3D object detection and BEV map segmentation, which can maintain resilient performance under arbitrary sensor input. Plenty of real-world sensor corruptions are formulated as well as methodically experimented with and analyzed to verify its robustness.
\item \model\ involves the \moe{ } structures to alleviate tasks conflict when performing 3D detection and segmentation tasks with the same trained weights.
\item \model{} achieves state-of-the-art performance on nuScenes dataset~\cite{caesar-20cvpr-nuscenes}. It's the first method designed for both sensor failures and tasks conflict. We hope \model\ will facilitate future research.
\end{enumerate}

\section{Related Work}

\subsection{LiDAR-Camera Fusion}
In 3D perception tasks, the effectiveness of multi-sensor fusion is notable. Therefore, researchers have focused on better combining geometric-centric point clouds and semantic-centric images. Existing methods mainly focus on three aspects: proposal-level~\cite{chen-17cvpr-mv3d}, point-level~\cite{chen-2022-futr3d,bai-22cvpr-transfusion,li-22arxiv-deepfusion}, and feature-level~\cite{liu-2022-bevfusion, liang-22cvpr-bevfusion}. While attempting to combine multiple modalities, existing approaches typically adopt similar fusion strategies, including query-indexing fusion~\cite{bai-22cvpr-transfusion} and channel-wise fusion~\cite{liu-2022-bevfusion,liang-22cvpr-bevfusion}. 
Though showing effective in full-modality input, the above fusion manners usually associate multi-modal features tightly among points or pixels, causing features in these methods to be susceptible to spatial misalignment issues and suffering from system collapses due to sensor missing. In our work, we explore an approach that can handle arbitrary modality freely to mitigate the above limitations.

\subsection{Sensors-Failure Perception} Sensor failures can significantly impact the accuracy of 3D perception, thereby jeopardizing the safety of autonomous driving. Therefore, conducting research on sensor failures is of utmost practical importance.  Prior arts have made preliminary attempts to propose robust frameworks that address specific sensor failure scenarios on a case-by-case basis. These scenarios include camera-views drop~\cite{bai-22cvpr-transfusion}, changes in illumination conditions~\cite{liu-2022-bevfusion,chen-2022-futr3d,bai-22cvpr-transfusion}, noisy inputs~\cite{li-22cvpr-deepfusion}, and etc. More recently, ~\cite{yu-22arxiv-benchmarking} provided a comprehensive benchmark for verifying the robustness of methods mainly on sensor degradation.
While previous works mainly focus on sensor corruptions, sensor absence is rarely noticed. In this work, we propose a novel perception framework that analyzes not only the aspects of sensor corruptions but also sensor absence comprehensively.

\subsection{Multi-Task Learning}
Multi-task learning denotes performing multiple task predictions with one set of trained weights, which appeals to the research community for its practical values, such as the complementary performance, less computational cost, etc. However, multi-task learning can also be challenging, as the model must learn to balance various objectives of each task and avoid tasks conflict. In 3D perception, the tasks conflict has been proven by  M\({}^{\textbf{2}}\)BEV~\cite{Xie-22arxiv-m2bev} and BEVFormer~\cite{li-22eccv-bevformer}, which both perform joint training on camera-only 3D object detection and BEV map segmentation. It is empirically found that joint training leads to severe accuracy degradation on each single task.
Prior arts do not address the issues effectively due to their unified models. Inspired by Mixture-of-Expert (MoE), which is specially designed for large language model in the Natural Language Processing~\cite{fedus-21corr-switchT, Lepikhin-21iclr-gshard, Jaszczur-21nips-sparsemoe, Yang-19nips-condconv, Wang-19uai-deepmoe} and self-supervised field~\cite{chowdhery-2022-palm,ge-21nips-care,ge-23iclr-snclr,driess-2023-palm}, we introduce a robust fusion module with a new \moe{-FFN} layer. The main purpose is to mitigate the gradient conflict between detection and segmentation to achieve a more balanced performance. \model{} is the first framework introducing MoE into 3D object detection and BEV map segmentation as a multi-modal, multi-task, and robust approach.

\begin{figure*}[t]
    \centering
    \resizebox{\textwidth}{!}{
        \includegraphics[width=0.47\textwidth]{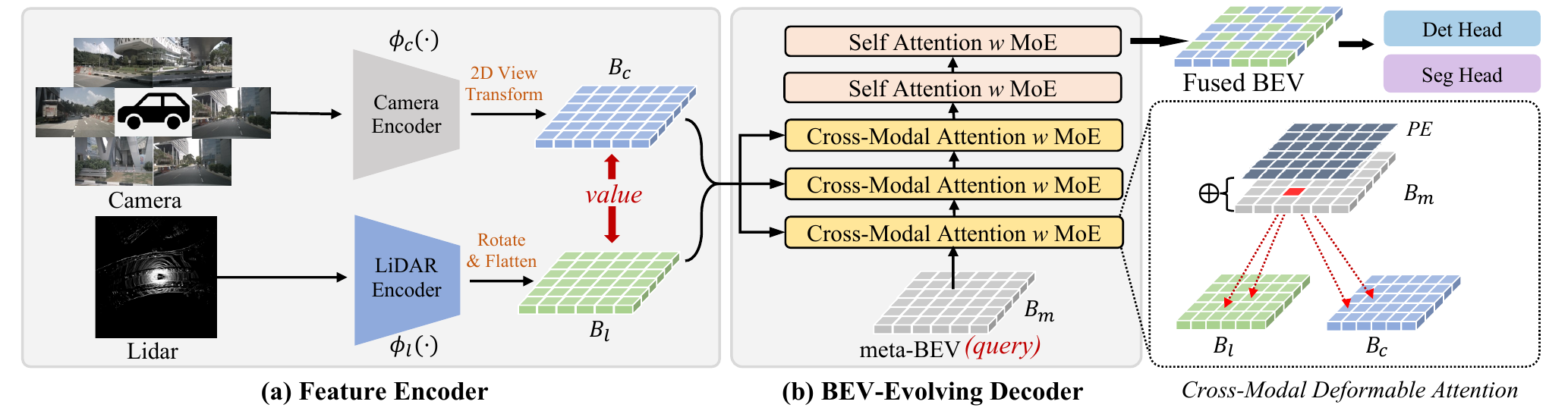}
    }
    \caption{\textbf{An overview of \model\ framework.} The multi-modal inputs are separately processed by the camera encoder $\phi_c(\cdot)$ and LiDAR encoder $\phi_l(\cdot)$ to produce the BEV representations $B_c, B_l$. To generate the fused BEV features, a BEV-Evolving decoder takes multi-modal BEV representations and an external initialized meta-BEV feature (as a query feature) for correlation computation. Task specific heads take the fused features for 3D detection.}
    \label{fig:pipeline}

\end{figure*}

\section{\model\ Method}
We introduce a new baseline, which targets solving a range of sensor failures on 3D object detection and BEV map segmentation tasks. As opposed to existing perception methods that heavily rely on the complete sensor inputs, we connect different modalities through a parameterized \emph{meta-BEV query} and perform cross-modal attention to integrate both semantic and geometry representations from cameras and LiDAR.
As shown in Figure~\ref{fig:pipeline}, the overall pipeline consists of a multi-modalities feature encoder, a BEV-Evolving decoder with cross-modal deformable attention, and task-specific heads. 
Architecture designs are detailed as follows.

\subsection{BEV Feature Encoder Overview}
\model\ generates fused features in the BEV space, as BEV representation provides a solution to combine image features with insufficient geometric knowledge and LiDAR features lacking semantic understanding in a unified domain, allowing for complementary fusion. Additionally, the regularity of the BEV feature facilitates the effective integration of various advanced task heads, which could benefit plenty of perception tasks.

\noindent{\textbf{Camera/LiDAR to BEV.}}
We build our multi-modal feature encoder based on the state-of-the-art perception method BEVFusion~\cite{liu-2022-bevfusion}, which takes multi-view-images and LiDAR-points pair as input and transforms the camera features into BEV space with depth prediction and geometric projection, respectively. 
Specifically, we adopt a camera backbone $\phi_c(\cdot)$ to generate $\mathit{N}$-view 2D image representations $F_c^{i}, i \in [1,2,\cdots, N]$. Then, following LSS~\cite{philion-20eccv-lss}, we scatter the dense feature $F_c^{i}$ into discrete 3D space and compress the generated 3D voxel features $V_c \in \mathbb{R}^{D\times C\times X\times Y}$ into the camera BEV representations $B_c \in \mathbb{R}^{C \times X \times Y}$ with the pillar format~\cite{liu-2022-bevfusion}, where $\mathit{D}, \mathit{C}, (\mathit{X}, \mathit{Y})$ denote the depth, dimension and spatial resolution of BEV feature, respectively. As for the LiDAR points, voxelization and sparse 3D convolutions $\phi_l(\cdot)$ are utilized for encoding LiDAR BEV representation $B_l$.

Subsequently, Previous works~\cite{liu-2022-bevfusion,liang-22cvpr-bevfusion} usually involve concatenating and CNN-based feature fusion, which may cause significant performance drop under sensor corruptions or even system collapse when sensor failures occur. As mentioned in Sec.~\ref{sec:intro}, CNNs exhibit limited effectiveness in large-corruption of input modality (due to the local-aggregation characteristic) and are inappropriate to deal with the absent input modality caused by sensor missing (due to the weight-sharing modeling). We solve the above challenges by introducing \textit{BEV-Evolving Decoder} as follows.

\subsection{BEV-Evolving Decoder} \label{sec:bev-evolving decoder}
The BEV-Evolving decoder consists of three key components: the cross-modal attention layers, the self-attention layers, and the plug-and-play \moe\ blocks. The above two components facilitate fusing arbitrary modalities, while the \moe\ blocks are designed for mitigating the tasks conflict.

\begin{figure*}[t]
\begin{center}

    \resizebox{\textwidth}{!}{
        \includegraphics[width=0.47\textwidth]{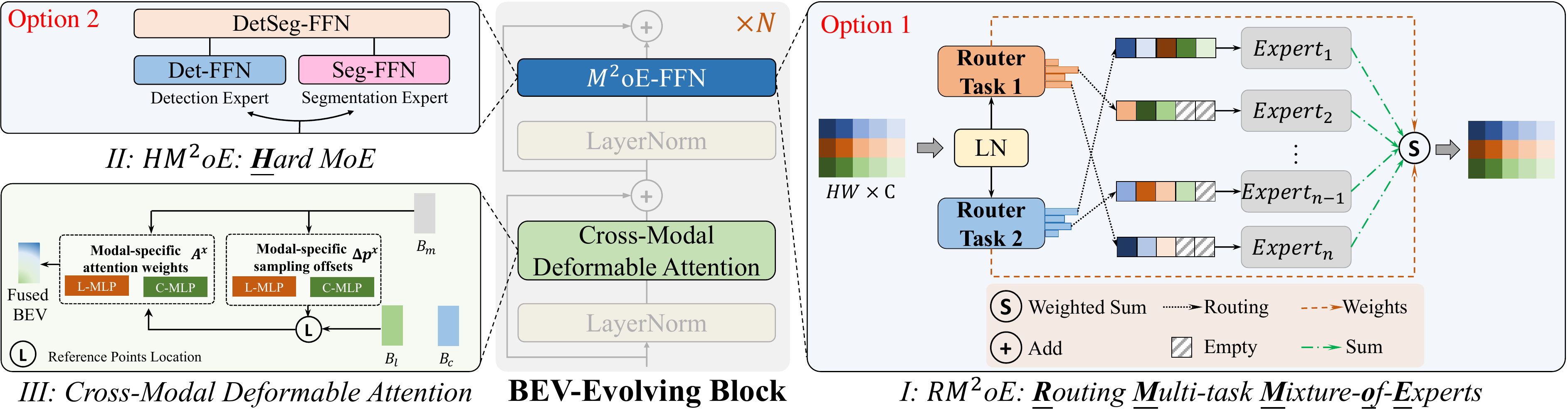}
    }
    \caption{\textbf{Detailed illustration of a BEV-Evolving block.} In the cross-modal deformable attention layer, We incorporate model-specific MLP layers to enable flexibly calculating sampling offsets and attention weights for an arbitrary modality. In FFN layer, we incorporate two MoE options to alleviate tasks conflict in multi-task learning. 
    } \label{fig:evolving_block}
\end{center}
\vspace{-2em}
\end{figure*}

\noindent{\textbf{Cross-Modal Attention Layer.}} We first initialize a set of dense BEV queries, each representing the feature within a specific spatial grid, termed meta-BEV ($B_m$). $B_m$ is added with position embeddings (PE) and then correlates with either a camera BEV feature ($B_c$), LiDAR BEV feature ($B_l$), or both. To improve the computational efficiency, we adopt deformable attention $\operatorname{DAttn}(\cdot)$ ~\cite{zhu-2020-deformable}. However, the original implementation $\operatorname{DAttn}(\cdot)$ is not suitable for processing arbitrary inputs, as it utilizes one unified MLP layer to sample reference points $\Delta p$ and attention weights $\mathit{A}$. 
We then introduce model-specific MLP layers (i.e., C-MLP for cameras and L-MLP for LiDAR in Fig.~\ref{fig:evolving_block} III) for the flexibility of fusion. 
Given any input BEV representations $x\in [B_c, B_l]$ as value features, we first generate model-specific sampling offsets $\Delta p^x$ and attention weights $\mathit{A}^x$ from query $B_m$. We then combine the pixel coordinates of $B_m$ and the corresponding offsets $\Delta p^x$ to locate the sampled value features. After re-scaling the sampled features by the attention weights $\mathit{A}^x$, meta-BEV is updated from informative sensor features. The overall process is formulated in the following,
\vspace{-2mm}
\begin{equation} \label{eq:datt} 
\begin{aligned}
    & \mathrm{DAttn}(B_m,p,x) = \\
    &  \sum_{m=1}^{M} W_m [\sum_{x\in[B_c\cup B_l]} \sum_{k=1}^{K} A_{mk}^x  \cdot {W_{m}^{'}}x (p+ \Delta p_{mk}^{x})],
\end{aligned}
\end{equation} 
where $m$ denotes the attention head, $K$ denotes the number of sampled keys, and $p$ denotes reference points. We use $W_{m}$ and $W_{m}^{'}$ to represent the learnable projection matrixes. 
Note that in Eq.~\ref{eq:datt}, the input feature $\mathit{x}$ could be either $B_c$, $B_l$, or both. It enables \model\ to flexibly utilize multi-modal features for the deformable attention calculation during both training and testing.

The cross-attention mechanism performs the fusion process layer by layer, enabling the meta-BEV to iteratively ``evolve'' into fused features that capture both semantic and geometric information from the cameras and LiDAR modalities, respectively.
Note that we sequentially incorporate $\mathit{N}$ cross-modal attention layers to capture inter-correlations among the meta-BEV and diverse sensor features. Nonetheless, we have not yet explicitly modeled the intra-correlations, which refer to as the connection among different queries. Therefore, we introduce the self-attention layers to facilitate fused features capturing the intra-correlations. The details are discussed below.

\noindent{\textbf{Self-Attention Layer.}} To construct the self-attention layer, we downgrade the modal-specific MLP into the unified MLP layer. Besides, to sufficiently model intra-correlations among queries, the input feature $\mathit{x}$ in Eq.~\ref{eq:datt} is substituted by $B_m$, resulting in a self-attention calculation $\mathrm{DAttn}(B_m, p, B_m)$. We assemble only \emph{M=2} self-attention layers in the BEV-Evolving Decoder, achieving a favorable trade-off between performance and computational efficiency. By modeling inter-modalities and intra-queries correlations, the fused BEV features are finally output for 3D predictions. It's experimentally found that the hybrid designs, combining both intra- and inter-correlations, provide comprehensive modeling on the fused BEV features, thus benefiting various tasks (as validated in Sec.~\ref{sec:abla}).

\noindent{\textbf{\moe{} Block.}} As shown in Fig.~\ref{fig:evolving_block} I and II, following the previous setting of \cite{shazeer-17iclr-outrageously} that is designed for large language modeling via a mixture of experts layers (MoE), we incorporate the MLP layer in our BEV-Evolving block with MoE to introduce our \moe{} blocks for multi-task learning. We first illustrate the R\moe (as shown in Fig.~\ref{fig:evolving_block} I) with the following formulation,
\begin{equation} \label{eq:m2oe}
\mathrm{M^2}\mathrm{oE}(x)= \sum_{i=1}^{t} \mathcal{R}{(x)_i} \mathcal{E}_{i}(x), t \ll E
\vspace{-2mm}
\end{equation}
where $x \in \mathbb{R}^{D}$ is the input tokens to the R\moe-FFN layer, $\mathcal{R}: \mathbb{R}^{D} \rightarrow \mathbb{R}^{E}$ is a routing function which assigns each token to its belonging experts and $\mathcal{E}_{i}: \mathbb{R}^{D} \rightarrow \mathbb{R}^{D}$ is the token processing in expert $i$. Both $\mathcal{E}_{i}$ and $\mathcal{R}$ are implemented by MLP layers, and $E$ is a hyper-parameter as the total expert capacity. With sparse $\mathcal{R}$ where top $t$ probability is selected, each token can only be routed to $t \ll E$ experts, while a lot experts remain inactive. Under the extreme condition of $t=1$, the additional calculation only comes from $\mathcal{R}$. For H\moe, Fig.~\ref{fig:evolving_block} II depicts a degenerate version when $E=task\ number$ and $\mathcal{R}=1$. In this scenario, tokens bypass the router allocation process and pass through the task-specific FFN networks before being merged by the task-fusion network. This super-linear scaling of the MLP layer facilitates multi-task training and inference by enabling the router to select appropriate experts. During network optimization, such a process mitigates tasks conflict between 3D object detection and BEV map segmentation by separating the conflicting gradients (caused by diverse training objectives) with different experts.

\subsection{Sensor Failures}
A practical perception model is required to perform effectively even encountering corrupted or absent inputs. To this end, we define a series of sensor failures to simulate both sensor corruptions and complete sensor absence. For sensor corruptions, we include six types: 1) \textit{Limited Field of LiDAR (LF)}, which occurs when LiDAR data can only be collected from a portion of the field of view due to incorrect collection or partial hardware damage~\cite{yu-22arxiv-benchmarking}; 2) \textit{Missing of Objects (MO)}, which appears when certain materials prevent some LiDAR points from being reflected~\cite{yu-22arxiv-benchmarking}; 3) \textit{Beams Reduction (BR)}, occurring due to limited power supply or sensor processing capabilities; 4) \textit{View Drop (VD)} and 5) \textit{View Noise (VN)}, resulting from camera faults; and 6) \textit{Obstacle Occlusion (OO)}, that is a real-world phenomenon where objects are occluded from the cameras. Furthermore, we evaluate \model\ using extreme sensor absence scenarios, including \textit{ Missing Camera(MC)} and \textit{Missing LiDAR(ML)}. As such, our evaluation takes into account both sensor corruptions and absence.

\subsection{Switched Modality Training} \label{sec:switch}
The unique designs (i.e., modal-specific modules) in the BEV-Evolving block enable \model\ flexibly processing of either camera features, LiDAR features, or both. We propose a switched-modality training scheme to ensure precise predictions by using arbitrary modalities. This alternating strategy simulates real-world conditions during training, as \model\ randomly receives inputs from the aforementioned modalities with predetermined probability. As a result, \model\ can conduct inferences on any input modality, thereby increasing its practicality in autonomous driving. Importantly, this approach requires only one set of pre-trained weights for all the model's deployment.

\begin{table*}[t]
\begin{center}
\resizebox{\linewidth}{!}{
    \begin{tabular}{lc | c| cc |ccccccc}
        \toprule
        Methods &  Modality & MTL & mAP(val) & NDS(val) & Drivable & Ped.Cross & Walkway & Stop Line & Carpark & Divider & Mean\\
        \midrule
        M\({}^{\mbox{2}}\)BEV\cite{Xie-22arxiv-m2bev}   & C & \xmark  & 41.7 & 47.0     & 77.2 & -    & -    & -    & -    & 40.5 & - \\
        BEVFormer\cite{li-22eccv-bevformer}             & C & \xmark & 41.6 & 51.7      & 80.1 & - & - & - & - & 25.7 & - \\

        BEVFusion\cite{liu-2022-bevfusion}              & C & \xmark & 35.6 & 41.2      & 81.7 & 54.8 & 58.4 & 47.4 & 50.7 & 46.4 & 56.6\\
        X-Align\cite{Borse-23wacv-xalign}               & C & \xmark & - & -            & 82.4 & 55.6 & 59.3 & 49.6 & 53.8 & 47.4 & 58.0 \\

        \rowcolor{gray!15}
        \textbf{\model-T} & & & 49.4 & 49.7 & & & & & & & \\
        \rowcolor{gray!15}
        \textbf{\model-C}  & \multirow{-2}{*}{C} & \multirow{-2}{*}{\xmark} & \textbf{55.5} & \textbf{60.4} & \multirow{-2}{*}{\textbf{83.3}} & \multirow{-2}{*}{\textbf{56.7}} & \multirow{-2}{*}{\textbf{61.4}} & \multirow{-2}{*}{\textbf{50.8}} & \multirow{-2}{*}{\textbf{55.5}} & \multirow{-2}{*}{\textbf{48.0}} & \multirow{-2}{*}{\textbf{59.3}} \\
        \midrule
        PointPillars\cite{lang-22cvpr-pointpillar}      & L & \xmark & 52.3 & 61.3       & 72.0 & 43.1 & 53.1 & 29.7 & 27.7 & 37.5 & 43.8\\
        CenterPoint\cite{yin-21cvpr-centerpoint}        & L & \xmark & 59.6 & 66.8       & 75.6 & 48.4 & 57.5 & 36.5 & 31.7 & 41.9 & 48.6\\
        BEVFusion\cite{liu-2022-bevfusion}              & L & \xmark & 64.7 & 69.3       & 75.6 & 48.4 & 57.5 & 36.4 & 31.7 & 41.9 & 48.6\\

        \rowcolor{gray!15}
        \textbf{\model-C} &&& 62.5 & 68.6 &&&&&&& \\
        \rowcolor{gray!15}
        \textbf{\model-T} & \multirow{-2}{*}{L} & \multirow{-2}{*}{\xmark} & 64.2 & \textbf{69.3} & \multirow{-2}{*}{\textbf{87.9}} & \multirow{-2}{*}{\textbf{63.4}} & \multirow{-2}{*}{\textbf{71.6}} & \multirow{-2}{*}{\textbf{55.0}} & \multirow{-2}{*}{\textbf{55.1}} & \multirow{-2}{*}{\textbf{55.7}} & \multirow{-2}{*}{\textbf{64.8}} \\
        \midrule
        PointPainting\cite{vora-20cvpr-pointpainting}   & L+C & \xmark & 65.8 & 69.6     & 75.9 & 48.5 & 57.1 & 36.9 & 34.5 & 41.9 & 49.1\\
        MVP\cite{yin-21nips-mvp}                        & L+C & \xmark & 66.1 & 70.0     & 76.1 & 48.7 & 57.0 & 36.9 & 33.0 & 42.2 & 49.0\\
        TransFusion\cite{bai-22cvpr-transfusion}        & L+C & \xmark & 67.3 & 71.2     & - & - & - & - & - & - & - \\
        BEVFusion\cite{liu-2022-bevfusion}              & L+C & \xmark & 68.5 & 71.4     & 85.5 & 60.5 & 67.6 & 52.0 & 57.0 & 53.7 & 62.7\\
        X-Align\cite{Borse-23wacv-xalign}              & L+C & \xmark & - & - & 86.8 & 65.2 & 70.0 & 58.3 & 57.1 & 58.2 & 65.7 \\
        \rowcolor{gray!15}
        \textbf{\model-T}   & L+C & \xmark & 68.0 & \textbf{71.5}  & \textbf{89.6} & \textbf{68.4} & \textbf{74.8} & \textbf{63.3} & \textbf{64.4} & \textbf{61.8} & \textbf{70.4}\\
        \hline
        \midrule
        BEVFusion$^\dagger$\cite{liu-2022-bevfusion}          & L+C & \cmark & - & 69.7 & - & - & - & - & - & - & 54.0 \\
        BEVFusion$^\ddagger$\cite{liu-2022-bevfusion}          & L+C & \cmark & 65.8 & 69.8 & 83.9 & 55.7 & 63.8 & 43.4 & 54.8 & 49.6 & 58.5\\
        \rowcolor{gray!15}
        \textbf{\model-MTL}$^\dagger$      & L+C & \cmark & 65.6 & 69.5 & \textbf{88.7} & \textbf{64.8} & \textbf{71.5} & \textbf{56.1} & \textbf{58.7} & \textbf{58.1} & \textbf{66.3} \\
        \rowcolor{gray!15}
        \textbf{\model-MTL}$^\ddagger$      & L+C & \cmark & 65.4 & \textbf{69.8} & \textbf{88.5} & \textbf{64.9} & \textbf{71.8} & \textbf{56.7} & \textbf{61.1} & \textbf{58.2} & \textbf{66.9} \\
        \bottomrule
    \end{tabular}
}
\end{center}

\caption{\textbf{Comparisons with SoTA methods on nuScenes val set.} We use -C and -T to denote equipping \model\ with the CenterPoint head~\cite{yin-21cvpr-centerpoint} and Transfusion head~\cite{bai-22cvpr-transfusion}. MTL stands for testing multi-tasks with the same model. $\dagger$ and $\ddagger$ stand for separating or sharing the BEV feature encoder, respectively. MetaBEV outperforms the SoTA multi-modal fusion methods by +4.7\% mIOU on nuScenes(val) BEV map segmentation and achieves comparable 3D object detection performance. \model\ also performs best in multi-task learning.}
\label{tab:nuscenes_det_val}

\end{table*}

\section{Experiments}
In this section, we detail the implementations and experimental settings on both sensor failures and tasks conflict. The performances on 3D detection and BEV map segmentation are presented to validate the effectiveness, flexibility, and robustness of our \model.

\subsection{Implementation Details} \label{sec:implementation}
{\flushleft \bf Network Architectures.} Swin-T~\cite{liu-21iccv-swin} and VoxelNet~\cite{zhou-18cvpr-voxelnet} are adopted as feature encoders for the cameras and LiDAR, respectively. In the BEV-Evolving decoder, we employ four cross-modal attention layers and two self-attention layers to produce a fused BEV. We initialize the meta-BEV with a resolution of $180\times 180$ to capture fine-grained correlations. Subsequently, we apply an FPN layer~\cite{lin-17cvpr-fpn} to generate multi-scale features from the fused BEV. Unless specified otherwise, we use a Transformer head for 3D detection~\cite{bai-22cvpr-transfusion,carion-20eccv-detr,wang-2022-detr3d} and a CNN head~\cite{liu-2022-bevfusion} for map segmentation.

{\flushleft \bf Datasets and Evaluation Metrics.} We evaluate \model\ on nuScenes~\cite{caesar-20cvpr-nuscenes}, a large-scale multi-modal dataset for 3D detection and map segmentation. The dataset is split into 700/150/150 scenes for training/validation/testing. It contains data from multiple sensors, including six cameras, one LiDAR, and five radars. For camera inputs, each frame consists of six views of the surrounding environment at one specific timestamp. We resize the input views to $256\times704$ resolution and voxelize the point cloud to 0.075m and 0.1m for detection and segmentation, respectively.
Our evaluation metrics align with~\cite{caesar-20cvpr-nuscenes}. For 3D detection, we utilize the standard nuScenes Detection Score (NDS) and mean Average Precision (mAP). For BEV map segmentation, we follow~\cite{liu-2022-bevfusion} to calculate the Mean Intersection over Union (mIoU) for map segmentation on the overall six categories.

{\flushleft \bf Training configurations.} We follow the image and LiDAR data augmentation strategies from MMDetection3D~\cite{mmdet3d2020} to enlarge the training samples' diversities.  AdamW~\cite{loshchilov-2017-adamw} is utilized with a weight decay of 0.05 and a cyclical learning rate schedule~\cite{smith-17wacv-cyclic} for optimization. We take overall 26 training epochs for 3D detection, and 20 epochs for BEV map segmentation, with CBGS~\cite{zhu-2019-cbgs} to balance the data sampling. \model\ is trained on 8 A100 GPUs. For the switched-modality training, the training ratios for camera/LiDAR/both are uniformly set to be 1/3, 1/3, and 1/3, respectively. Further details are available in the appendix.

\begin{table*}[t]
\footnotesize
\begin{center}
    \renewcommand{\tabcolsep}{5mm}
    \begin{tabular}{c | ccc  ccc  ccc }
    \toprule
        \multirow{2}{*}{Methods} & \multicolumn{3}{c}{Camera + Lidar} & \multicolumn{3}{c}{Missing Camera} & \multicolumn{3}{c}{Missing LiDAR} \\
        \cmidrule[0.2mm](r){2-4}\cmidrule[0.2mm](r){5-7}\cmidrule[0.2mm](r){8-10}
        & mAP & NDS & mIoU & mAP & NDS & mIoU & mAP & NDS & mIoU \\
    \midrule
        TransFusion~\cite{bai-22cvpr-transfusion} & 67.3 & 71.2 & -- & 61.6 & 67.4 &-- &-- &-- &--  \\
        BEVFusion~\cite{liu-2022-bevfusion} & 68.5 & 71.4 & 62.7 & \gain{61.8} & \gain{67.5} & \gain{4.1} & \gain{0.5} & \gain{7.1} & \gain{36.7} \\
    \rowcolor{gray!15}        
        MetaBEV & 68.0 & 71.5 &  70.4 & 63.6 & 69.2 & 53.7 & 39.0 & 42.6 & 54.4 \\
    \bottomrule
    \end{tabular}
\end{center}
\caption{\textbf{Experimental comparisons on extreme sensor missing.} \model\ is able to totally drop the features from the missing modalities for inference, while others cannot. We attempt to replace the missing features with zero in other works so that they can output results, which are colored as \gain{blue}.
\model\  still consistently outperforms prior works when facing extreme sensor absence.}
 \label{tab:missing_modal}
\vspace{-2mm}
\end{table*}

\begin{table*}[t]
\begin{center}
\resizebox{\textwidth}{!}{
    \begin{tabular}{c |c | cc  cc  cc  cc  cc cc}
    \toprule
        \multirow{3}{*}{Methods} & \multirow{3}{*}{Evaluation} & \multicolumn{2}{c}{Limited Field} & \multicolumn{2}{c}{Missing Objects} & \multicolumn{2}{c}{Beam Reduction} & \multicolumn{2}{c}{View Drop} & \multicolumn{2}{c}{View Noise} & \multicolumn{2}{c}{Obstacle Occlusion } \\
        & & \multicolumn{2}{c}{\gain{[-60,60]}} & \multicolumn{2}{c}{\gain{rate=1.0}} & \multicolumn{2}{c}{\gain{4 beams}} & \multicolumn{2}{c}{\gain{6 drops}} & \multicolumn{2}{c}{\gain{6 noise}} & \multicolumn{2}{c}{\gain{w occlusion}} \\
        \cmidrule[0.2mm](r){3-4}\cmidrule[0.2mm](r){5-6}\cmidrule[0.2mm](r){7-8}\cmidrule[0.2mm](r){9-10}\cmidrule[0.2mm](r){11-12}\cmidrule[0.2mm](r){13-14}
        & & NDS & mIoU & NDS & mIoU & NDS & mIoU & NDS & mIoU & NDS & mIoU & NDS & mIoU \\
    \midrule
        BEVFusion~\cite{liu-2022-bevfusion} &  \multirow{2}{*}{zero-shot} & 41.6 & 47.2 & 62.1 & 61.8 & 58.3 & 55.3 & 67.5 & 4.1 & 66.9 & 25.8 & 68.6 & 45.8 \\     
        MetaBEV & & 47.0 & 61.2 & 62.5 & 69.2 & 57.7 & 64.3 & 68.3 & 43.5 & 68.0 & 45.2 & 70.0 & 61.2 \\
    \midrule
        BEVFusion~\cite{liu-2022-bevfusion} &  \multirow{2}{*}{in-domain} & 42.6 & 55.5 & 65.2 & 61.9 & 51.2 & 59.0 & 68.3 & 52.8 & 68.3 & 53.1 & 69.7 & 59.3 \\     
        MetaBEV & & 54.3 & 62.1 & 68.5 & 69.9 & 54.6 & 66.4 & 69.3 & 67.9 & 69.2 & 67.5 & 70.3 & 70.2 \\
    \bottomrule
    \end{tabular}
    }
\end{center}
\caption{\textbf{Experimental comparisons on sensor corruptions.} Texts in \gain{blue} denote the corruption degree. More detailed results are available in the appendix. \model\ consistently outperforms BEVFusion on various sensor corruptions in zero-shot and in-domain tests.}
\label{tab:corruptions}
    \vspace{-1em}
\end{table*}

\subsection{Performance on full modalities}
We evaluate \model\ on 3D object detection and BEV map segmentation, utilizing either single or full modality inputs.
For 3D object detection, results in Tab.~\ref{tab:nuscenes_det_val} show that \model\ achieves superior performance with the camera modality, and gets comparable performance to the state-of-the-art  with both LiDAR and multi-modalities. For example, with CenterPoint-3D~\cite{yin-21cvpr-centerpoint}, \model-C achieves an impressive 60.4\% NDS on nuScenes val set with camera inputs, for LiDAR and multi-modal inputs, \model\ gains 69.3\% and 71.5\%, respectively, which is comparable to several SoTA methods.
Moreover, it is experimentally found that \model\ is effective in capturing fine-grained features for dense predictions, \eg, semantic prediction. For single-modality inputs, with only several MLP heads for prediction, \model\ achieves a 59.3\% mIoU on cameras, and an impressive 64.8\% mIoU on LiDAR (+16.2\% higher than BEVFusion~\cite{liu-2022-bevfusion}). Moreover, results in Tab.~\ref{tab:nuscenes_det_val} show we set a new state-of-the-art (SOTA) performance of 70.4\% mIoU on BEV map segmentation, outperforming the previous best model~\cite{Borse-23wacv-xalign} with +4.7\% mIoU and the second-best method~\cite{liu-2022-bevfusion} by +7.7\% mIoU.


\begin{table}[t]
\begin{center}
\renewcommand{\tabcolsep}{3mm}
    \resizebox{\linewidth}{!}{
    \begin{tabular}{cc | c c | c}
        \toprule
        && mAP & NDS & mIoU \\
        \midrule
        \multicolumn{2}{l|}{Detection only}          & 67.6  & 71.0  & - \\
        \multicolumn{2}{l|}{Segmentation only}       & - & - & 70.4 \\
        \midrule
        H-MoE & R-MoE & & & \\
        \midrule
        \xmark & \xmark                             & 64.8  & 69.4  & 64.7 \\
        \cmark & \xmark                             & \textbf{65.6}~\gain{(+0.8)}  & 69.5~\gain{(+0.1)}  & 66.3~\gain{(+1.6)} \\
        \xmark & \cmark                             & 65.4~\gain{(+0.6)}  & \textbf{69.8}~\gain{(+0.4)}  & \textbf{66.9}~\gain{(+2.2)} \\
        \bottomrule
    \end{tabular}
    }
\end{center}
\caption{\textbf{We experiment on two kinds of MoE structures for Multi-Task Learning}, and both show significant advantages.}
\label{tab:mtl_exp}
\end{table}

\begin{table*}[t]
\centering
\begin{subtable}[t]{0.3\textwidth}
\begin{center}
    \begin{tabular}{l c}
    \toprule
     Layers   & mAP/DNS  \\
    \midrule
        4 cross & 61.7/67.0 \\
        6 cross & 63.0/68.3 \\
        8 cross & 62.5/68.2 \\
        \rowcolor{gray!15}
        \textbf{2 self+4 cross} & \textbf{64.2/69.3} \\
    \bottomrule
    \end{tabular}
    \caption{The number of self/Cross-layer}
    \label{tab:ablations-layernum-a}
\end{center}
\end{subtable}
\begin{subtable}[t]{0.25\textwidth}
\begin{center}
    \begin{tabular}{l c}
    \toprule
     Points   & mAP/DNS  \\
    \midrule
        4 ~~~~~~~& 62.3/67.9 \\
        \rowcolor{gray!15}
        \textbf{8} & \textbf{62.7/68.2} \\
        12  & 62.3/68.0 \\
        16  & 62.3/68.1 \\
    \bottomrule
    \end{tabular}  
    \caption{The number of reference points}
    \label{tab:ablations-rfp-b}
\end{center}
\end{subtable}
\begin{subtable}[t]{0.35\textwidth}
\begin{center}
    \begin{tabular}{l c c c c}
    \toprule
    Experts    & mAP/DNS  & mIoU \\
    \midrule
        \rowcolor{gray!15}
        1/2 Expert & 65.8/69.8 & 63.3 \\
        1/4 Expert & 64.9/69.5 & 65.9 \\
        1/8 Expert & 65.4/69.7 & 66.4 \\
        \rowcolor{gray!15}
        \textbf{2/8 Expert} & \textbf{65.4/69.8} & \textbf{66.9} \\
    \bottomrule
    \end{tabular}
    \caption{Expert number}
    \label{tab:ablations-expertnum-c}
\end{center}
\end{subtable}

\begin{subtable}[b]{1.0\textwidth}
\centering
    \resizebox{1.0\textwidth}{!}{
    \begin{tabular}{c|c|c c c|c c c}
        \toprule
        \multirow{2}{*}{Task} & \multirow{2}{*}{Metrics} & \multicolumn{3}{c|}{Vanilla training} & \multicolumn{3}{c}{Switched Modality Training}  \\
        & & Missing LiDAR & Missing Camera & Full & Missing Lidar & Missing Camera & Full \\
        \midrule
        \multirow{2}{*}{3D detection} & NDS & 9.8 & 68.2 & 71.0 & 42.6~\gain{(+32.8)} & 69.2~\gain{(+1.0)} & 71.5~\gain{(+0.5)} \\
        & mAP & 2.9 & 62.4 & 67.4 & 39.0~\gain{(+36.1)} & 63.6~\gain{(+1.2)} & 68.0~\gain{(+0.6)} \\
        \midrule
        BEV-Map Segmentation & mIoU & 36.5 & 27.0 & 70.4 & 54.4~\gain{(+17.9)} & 53.7~\gain{(+16.7)} & 68.5~\textcolor{red}{(-1.9)} \\
        \bottomrule
    \end{tabular}
    }
    \caption{Evaluation of the NuScenes val dataset when Cameras or LiDAR totally fail. We test the performance with the switched modality
training on both 3D detection and BEV-map segmentation tasks.}
    \label{tab:ablations-switchtrain-d}
\end{subtable}
\caption{\textbf{Ablation studies for the model architectures and training strategy.} Default settings are marked in \colorbox{gray!15}{gray} in (a), (b), and (c). 
}
\label{tab:ablation}
\end{table*}

\subsection{Performance on corruption}
We proceed to evaluate \model\ under various sensor failure cases, including complete sensor missing and various sensor corruptions. For the sensor missing, we train \model\ with the switched modality scheme, as described in Sec.~\ref{sec:switch}. 
Previous fusion-based approaches, such as BEVFusion~\cite{liu-2022-bevfusion} and TransFusion~\cite{bai-22cvpr-transfusion}, heavily rely on multi-modal features and cannot explicitly handle the absence of features caused by sensor missing. Nevertheless, we attempt to replace missing modalities with zero-initialized features for them, allowing the prior methods for predictions.  Results in Tab.~\ref{tab:missing_modal} show that \model\ demonstrates a stronger anti-corruption ability. Specifically, in case of missing LiDAR, \model\ improves $+35.5\%$ detection NDS and $+17.7\%$ segmentation mAP upon BEVFusion~\cite{liu-2022-bevfusion}; when encountering missing cameras, \model\ surpasses BEVFusion~\cite{liu-2022-bevfusion} by +1.7\% NDS and 49.5\% mIoU.
Notably, even when the cameras are missing, \model\ still achieves 53.7\% mIoU, surpassing the SoTA LiDAR-Only methods by 5.1\% (i.e., CenterPoint~\cite{yin-21cvpr-centerpoint}) and 9.9\% (i.e., PointPillars~\cite{lang-22cvpr-pointpillar}); and when the LiDAR is missing, \model{} even outperforms the camera-specific BEVFusion by +1.4\% NDS (42.6\% \textit{v.s.} 41.2\%).

On the other hand, we conducted both zero-shot and in-domain tests on sensor corruptions. To evaluate the model's abilities in performing with unseen corrupted or missing data, we performed the zero-shot test by directly evaluating a trained model on such data. For the in-domain test, we train \model\ on corrupted data with random degrees, and then conduct evaluation on noise data with one specific degree. Tab.\ref{tab:corruptions} shows \model\ outperforms BEVFusion~\cite{liu-2022-bevfusion} on \emph{11/12 corruptions} on zero-shot evaluation. Moreover, when trained with randomly corrupted data, \model\ consistently outperformed BEVFusion. The results indicate that the fusion module, which is composed of CNNs (\eg, BEVFusion), has limited representational capabilities for corrupted data, while our approach leverages the attention mechanism for efficient modeling.

\subsection{Performance in Multi-Task Learning}
Tab.~\ref{tab:mtl_exp} presents the results of our multi-task learning (MTL) experiments for 3D object detection and BEV map segmentation. We begin with the framework designed for independent single-task learning with distinct task-specific heads and loaded a set of weights trained on multi-modal detection as pretrain. The model achieves 69.4\% NDS and 64.7\% mIoU for detection and segmentation, respectively, which is already a state-of-the-art performance. In light of this, we implement an MoE structure to help alleviate the gradient-conflict problem in MTL. Intuitively, we introduce a routing-based MoE structure, referred to as R\moe{}, to replace the vanilla MLP layer. This implementation results in a remarkable improvement of 0.4\% NDS in detection and 2.2\% mIoU in segmentation. Considering the additional computation required by the routing mechanisms, we propose to further experiment on a simplified version H\moe{} where the number of experts is adjusted to match that of the tasks, and the routing function is removed. We find it still achieves a notable improvement of 0.1\% NDS and 1.6\% mIoU. Upon the evaluation results, both implementations mitigate performance degradation in MTL to some extent.

\subsection{Ablation Studies} \label{sec:abla}
{\flushleft \bf Setup.} We present the ablation studies for analyzing the network architectures and training schemes. All the ablation studies are performed on nuScenes validation set, with the training configurations set as default in Sec.~\ref{sec:implementation}.

{\flushleft \bf Network Configurations.} We first explore the optimal architectures of the BEV-Evolving Decoder. Specifically, we analyze three components, including the combination of the layers, reference points number ($\mathit{p}$ in Eq.\ref{eq:datt}), and expert number ($\mathit{E}$ in Eq.~\ref{eq:m2oe}). Our findings in Tab.~\ref{tab:ablations-layernum-a} indicate that a few layers of cross-attention are sufficient to produce effective predictions, while stacking more cross-attention layers does not necessarily enhance the model's capacities (68.3\% and 68.2\% NDS for 6 and 8 cross-attention layers, respectively). The same findings also suit the reference points number, where increasing the sampling points does not necessarily lead to better performance, as shown in Tab.~\ref{tab:ablations-rfp-b}. 
On the contrary, we discovered that simply adding two layers of self-attention after the cross-attention layers significantly improve the prediction performance by +1.1\% NDS in Tab.~\ref{tab:ablations-layernum-a}. As mentioned in Sec.~\ref{sec:bev-evolving decoder}, the hybrid designs capture both intra- and inter-correlations simultaneously, resulting in complementary modeling. 
Moreover, in Tab.~\ref{tab:ablations-expertnum-c}, we discover that larger model capacity (\ie, incorporating more experts) can effectively alleviate the performance drop in 3D detection and map segmentation. Specifically, \model\ equipped with 2/8 expert (\ie activating the 2 experts with the highest scores from a total of 8 experts.) achieves the best performance compared to other configurations (66.9\% mIoU and 69.8\% NDS), while H\moe\ is unaffected due to the fixed number of experts being equivalent to the number of tasks.

{\flushleft \bf Switched Modality Training.}
For the training strategy, using switched modality scheme to simulate the sensors missing, though simple, is the key to enabling \model\ with robust performance on completely sensor-failure scenarios. In Tab.~\ref{tab:ablations-switchtrain-d}, comparing with the vanilla full-modality training approach, our proposed training strategy yields significant improvements on 3D detection by +32.8\% NDS and +1.0\% NDS, and improvements on map segmentation by +17.9\% mIoU and +16.7\% mIoU under LiDAR-missing and camera-missing scenarios, respectively. More intriguingly, we find that the switched modality training could even enhance the model performance on full-modalities, rising from 71.0\% NDS to 71.5\% NDS on 3D detection.

\vspace{-1mm}
\section{Conclusion}
In this paper, we present a novel framework, named \model{}, for the purpose of solving sensor failures in the bird’s-eye view (BEV) 3D detection and map segmentation. Our method integrates modal-specific layers into the cross-modal attention layer to enhance the fusion process, achieving appealing performance on full-modality inputs, and meanwhile \model{} effectively alleviates the significant performance degradation that is often caused by corrupted or missing sensor signals. We also introduce \moe{} to handle potential conflicts between tasks. We hope that \model{} will provide a more focused and effective approach for investigating sensor-failure scenarios in the autonomous driving field.

\vspace{-1mm}
{\flushleft \bf Limitations.} Though we adopted the deformable attention~\cite{zhu-2020-deformable} for efficiency, it unavoidably leads to a slight increase in network parameters compared to the lightweight solutions~\cite{liu-2022-bevfusion,liang-22cvpr-bevfusion}. Despite this, the benefits of mitigating sensor failures are highly desirable, and the additional computational overhead may be deemed acceptable. 
We acknowledge that lightweight networks could be an avenue for future research, potentially involving techniques such as network pruning, token reduction, etc.

\newpage
\vspace{3em}
{\centering
    {\Large \bf MetaBEV: Solving Sensor Failures for BEV  Detection and Map Segmentation} \\
    {\Large \bf $\mathrm{<}$Supplementary Material$\mathrm{>}$\\}
}

\vspace{2em}
We provide a detailed PyTorch-style pseudo-code to describe two key components in our \model, including the cross-modal attention and the \moe{} structure. Then we elaborate on the comprehensive sensor corruptions utilized in our evaluation, as well as the detailed training configurations employed for BEV 3D detection and map segmentation. Moreover, we present additional experiments and visualizations to further validate the effectiveness of \model{}.  We also provide \href{https://chongjiange.github.io/metabev.html}{more video demos} to illustrate how \model{} performs under various sensor failures.
Finally, we emphasize that the code and trained models used in our research will be made publicly available for future studies.

\renewcommand\thesection{\arabic{section}}

\section{Algorithms}
We present the PyTorch-style pseudo-code for two key components in \model, i.e., the cross-modal attention and \moe, in Algorithm~\ref{alg:attention} and Algorithm~\ref{alg:moe}, respectively.

\section{Implementation Details} \label{sec:supp_implementation}
\subsection{Datasets} We evaluate \model\ on nuScenes~\cite{caesar-20cvpr-nuscenes}, a large-scale multi-modal dataset for 3D detection and map segmentation, which contains high-resolution sensor data collected from diverse urban driving scenes, including a vast amount of annotated images, LiDAR scans, and 3D annotations. The dataset is generally split into 700/150/150 scenes for training/validation/testing. It contains data from multiple sensors, including six cameras, one LiDAR, and five radars. For camera inputs, each frame consists of six views of the surrounding environment at one specific timestamp. For LiDAR inputs, nuScenes collects points cloud reflected by the surroundings within 360 degrees. NuScenes is widely employed for evaluation on object detection, tracking, and semantic segmentation. In this work, we not only evaluate our model on complete input modalities from camera and LiDAR, but also conduct an in-depth investigation of the robustness of \model{} by utilizing pre-existing datasets to simulate and evaluate various corruptions. Details are illustrated as follows.

\begin{table*}[t]
\begin{center}
    \resizebox{1.0\textwidth}{!}{
    \begin{tabular}{c|c|c c c|c c c} 
        \toprule
        \multirow{2}{*}{Task} & \multirow{2}{*}{Epoch} & \multicolumn{3}{c|}{Vanilla training} & \multicolumn{3}{c}{Switched Modality Training}  \\
        & & Lidar-Missing & Camera-Missing & Full & Lidar-Missing & Camera-Missing & Full \\
        \midrule
        BEV-Map Segmentation & 20(vanilla)+8(SMT) & 36.5 & 27.0 & 70.4 & 54.4~\gain{(+17.9)} & 53.7~\gain{(+16.7)} & 68.5~\textcolor{red}{(-1.9)} \\
        BEV-Map Segmentation & 20(Vanilla)+20(SMT) & 36.5 & 27.0 & 70.4 & 63.4~\gain{(+26.9)} & 58.3~\gain{(+31.3)} & 69.2~\textcolor{red}{(-1.2)} \\
        \bottomrule
    \end{tabular} 
}
\end{center}

\vspace{2mm}
\caption{\textbf{Map segmentation results of Switched Modality Training with longer epochs}. Comparing the results of eight-epochs training in the main paper, we found that training for a longer schedule (e.g, 20 epochs) further improves \model's performance in map segmentation.
}
\label{tab:appendix_smt_seg}

\end{table*}

\subsection{Corruptions} For sensor corruptions, we introduce a total of six common corruptions, which are detailed in the following:

{\flushleft \bf 1. Limited Field (LF):} Nuscenes collects LiDAR point cloud data from a 360-degree perspective. However, due to hardware malfunctions, there may be certain angles where the point cloud data is missing, resulting in data collection that is less than 360 degrees. In this article, we follow~\cite{yu-22arxiv-benchmarking} to simulate missing LiDAR data at different angles. The cases include 360, 240, 180, 120 degrees. 

{\flushleft \bf 2. Missing of Objects (MO):} In the real world, differences in object color and surface material, among other factors, can result in a reduction in the number of reflection points detected by the LiDAR system. In this work, we simulate missing object points by probabilistically removing point cloud data from objects with a specific rate. The selected rates are set to 0.0, 0.1, 0.3, 0.5, 0.7, 1.0. 

{\flushleft \bf 3. Beam Reduction (BR):} Beam Reduction (BR) is a phenomenon that occurs in sensing systems where the available power supply or sensor processing capability is limited. In such systems, it is not possible to process the full set of data collected by the sensor, which leads to a reduction in the number of beams that are processed. In our work, we select 1,  4, 8, 16 and 32 beams for evaluation. 

{\flushleft \bf 4. View Noise (VN):}  It is a phenomenon whereby the captured view contains random variations or distortions that are not representative of the actual scene. VN can arise due to various factors, including sensor noise, electrical interference, atmospheric conditions, or compression artifacts. In this paper, we have adopted a strategy to evaluate the effectiveness of our approach by simulating VN through the replacement of 0 to 6 views with randomly generated noise. 

{\flushleft \bf 5. View Drop (VD):}  It refers to a real-world scenario where a portion of the visual scene is not captured by cameras, leading to a loss of information. VD can arise due to various factors such as incorrect camera positioning or hardware failure. In a similar manner to VN, we involve substituting zero-initialized inputs for up to six missing views during evaluation. 

{\flushleft \bf 6. Obstacle Occlusion (OO):} It refers to a perceptual phenomenon that occurs when objects within a scene are partially obscured from view by obstructions or occlusions. To simulate this phenomenon, we use a set of  predefined masks to perform alpha blending with the camera views, thereby generating an occluded view.

\subsection{Training Details} To train on the full-modalities, we adopt the following configurations.
To increase the diversity of training samples, we use standard image and LiDAR data augmentation strategies from MMDetection3D~\cite{mmdet3d2020}. We also utilize AdamW~\cite{loshchilov-2017-adamw} as our optimizer with a weight decay of 0.05 and a cyclical learning rate schedule~\cite{smith-17wacv-cyclic}. To balance the object classes during data sampling, we use CBGS~\cite{zhu-2019-cbgs}. Generally, we train for 26 epochs for 3D detection and 20 epochs for segmentation on 8 A100 GPUs to achieve the best results.  

\begin{figure*}[t]

\begin{center}
\footnotesize
\includegraphics[width=1\linewidth]{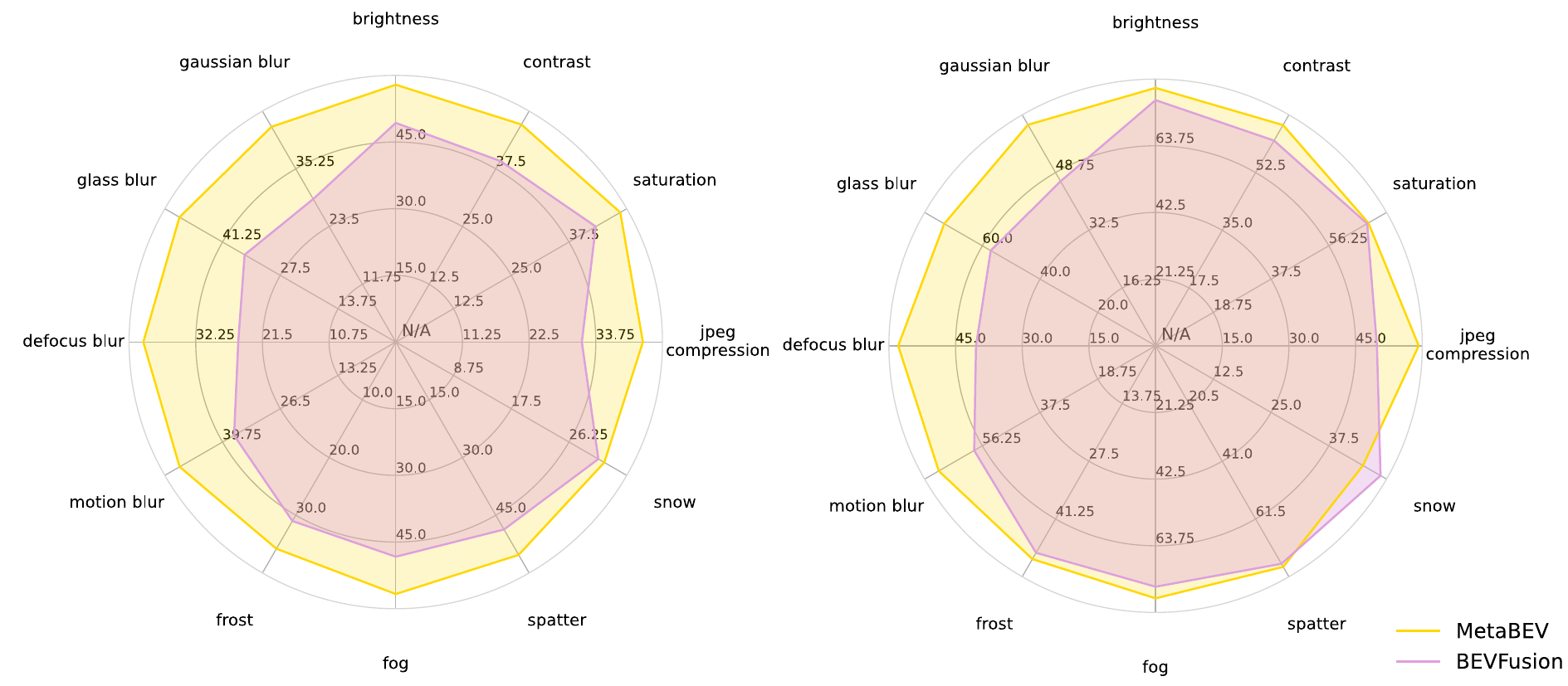}

\caption{\textbf{Map Segmentation results comparison under camera noises.} We show the map segmentation mIoU (\textbf{left}) and the corresponding retention (\textbf{right}) under different weather, blur and digital conditions to show the zero-shot performance in the real-world environment. Details could be found in Table~\ref{tab:noise_on_seg} and better view in color on screen.
} \label{fig:radar_mapSeg}
\end{center} 

\end{figure*} 

\begin{table*}[t]
\footnotesize
\begin{center}
    \resizebox{1.0\textwidth}{!}{
    \begin{tabular}{c | c | cccc | cccc | cccc }
    \toprule
        \multirow{2}{*}{Methods} & \multirow{2}{*}{Clear} & \multicolumn{4}{c|}{Blur} & \multicolumn{4}{c|}{Digital} & \multicolumn{4}{c}{Weather} \\
        \cmidrule[0.2mm](r){3-6}\cmidrule[0.2mm](r){7-10}\cmidrule[0.2mm](r){11-14}
        & & Motion & Defoc & Glass & Gauss & Bright & Contr & Satur & JPEG & Snow & Spatt & Fog & Frost \\
    \midrule
        \multirow{2}{*}{BEVFusion~\cite{liu-2022-bevfusion}} & \multirow{2}{*}{62.9} & 37.1 & 25.4 & 36.0 & 29.1 & 49.3 & 39.2 & 43.3 & 31.4 & 30.7 & 48.7 & 48.3 & 31.0 \\
        && (58.9\%) & (40.4\%) & (57.1\%) & (46.3\%) & (78.3\%) & (62.2\%) & (68.7\%) & (49.8\%) & \textbf{(48.7\%)} & (77.4\%) & (76.8\%) & (49.3\%) \\
    \midrule      
        \multirow{2}{*}{MetaBEV} & \multirow{2}{*}{\textbf{70.4}} & \textbf{49.6} & \textbf{40.7} & \textbf{51.5} & \textbf{43.8} & \textbf{57.9} & \textbf{47.1} & \textbf{48.6} & \textbf{41.7} & \textbf{31.6} & \textbf{55.3} & \textbf{56.7} & \textbf{35.8} \\
        && \textbf{(70.4\%)} & \textbf{(57.9\%)} & \textbf{(73.2\%)} & \textbf{(62.2\%)} & \textbf{(82.2\%)} & \textbf{(66.9\%)} & \textbf{(69.1\%)} & \textbf{(59.2\%)} & (44.9\%) & \textbf{(78.5\%)} & \textbf{(80.5\%)} & \textbf{(50.8\%)} \\ 
    \bottomrule
    \end{tabular}
    }
\end{center}
\vspace{-2mm}
\caption{\textbf{Experimental comparisons on map segmentation under different camera noises.}.}
 \label{tab:noise_on_seg}
\vspace{-2mm}
\end{table*}

For multi-task learning, we load the multi-modality model pre-trained on 3D detection task, and insert the segmentation head into the BEV-Evolving encoder. We then fine-tune the overall network with another 10 epochs. To balance multiple training objectives, we rescale the weights for detection and segmentation losses with 10 and 1, respectively and these hyper-parameters follow the setting in BEVFusion~\cite{liu-2022-bevfusion}.

For the switched-modality training, we begin by loading a task-specific pre-trained model. During each training iteration, we randomly input sensor information from either camera, LiDAR, or both in a proportion uniformly configured as 1/3, 1/3, 1/3.

\section{Additional Experiments}
In this chapter, Sec.~\ref{subsec:c1-result_smt} first demonstrates the performance of using Switched-Modality Training scheme to train \model\ on map segmentation for longer epochs (Table~\ref{tab:appendix_smt_seg}~\vs Table 5(d) in main paper). Secondly, Sec.~\ref{subsec:c2-corruptions_details} presents a more comprehensive analysis of the zero-shot and in-domain performance of \model\ on various corruption types compared to the results reported in Table 3. In addition, Sec.~\ref{subsec:c3-noise_mapseg} illustrates the effectiveness of our map segmentation approach in various real-world scenarios, like snow, fog scenes~\etc. Finally, Sec.~\ref{subsec:c4-visualizations} provides additional visualizations to facilitate further comparisons.

\subsection{Additional Results on Switch Modality Training} \label{subsec:c1-result_smt}

In our main paper, the results of map segmentation in Table 5(d) are achieved via training \model\ using the Switched-Modality training (SMT) strategy for 8 epochs. The epoch number here is set to be the same as the full modality fine-tuning schedule, and \model\ already demonstrates the satisfactory performance even when a single modality is absent (\eg, LiDAR or Camera). To further verify the performance of \model, we train it using SMT for total 20 epochs. The experimental results presented in Table~\ref{tab:appendix_smt_seg} demonstrate that our model's performance could be significantly improved on both single and full modalities with longer training schedule. For instance, as shown in Table~\ref{tab:appendix_smt_seg}, \model\ achieves a remarkable 63.4\% and 58.3\% mIoU on camera-only inputs and LiDAR-only inputs. These results consistently outperform the model trained with only 8 epochs by 7.0\% and 4.6\% mIoU, respectively.

\subsection{Detailed Comparisons on Sensor Corruptions} \label{subsec:c2-corruptions_details}

\begin{figure*}[t]
\begin{center}
\footnotesize
\includegraphics[width=1\linewidth]{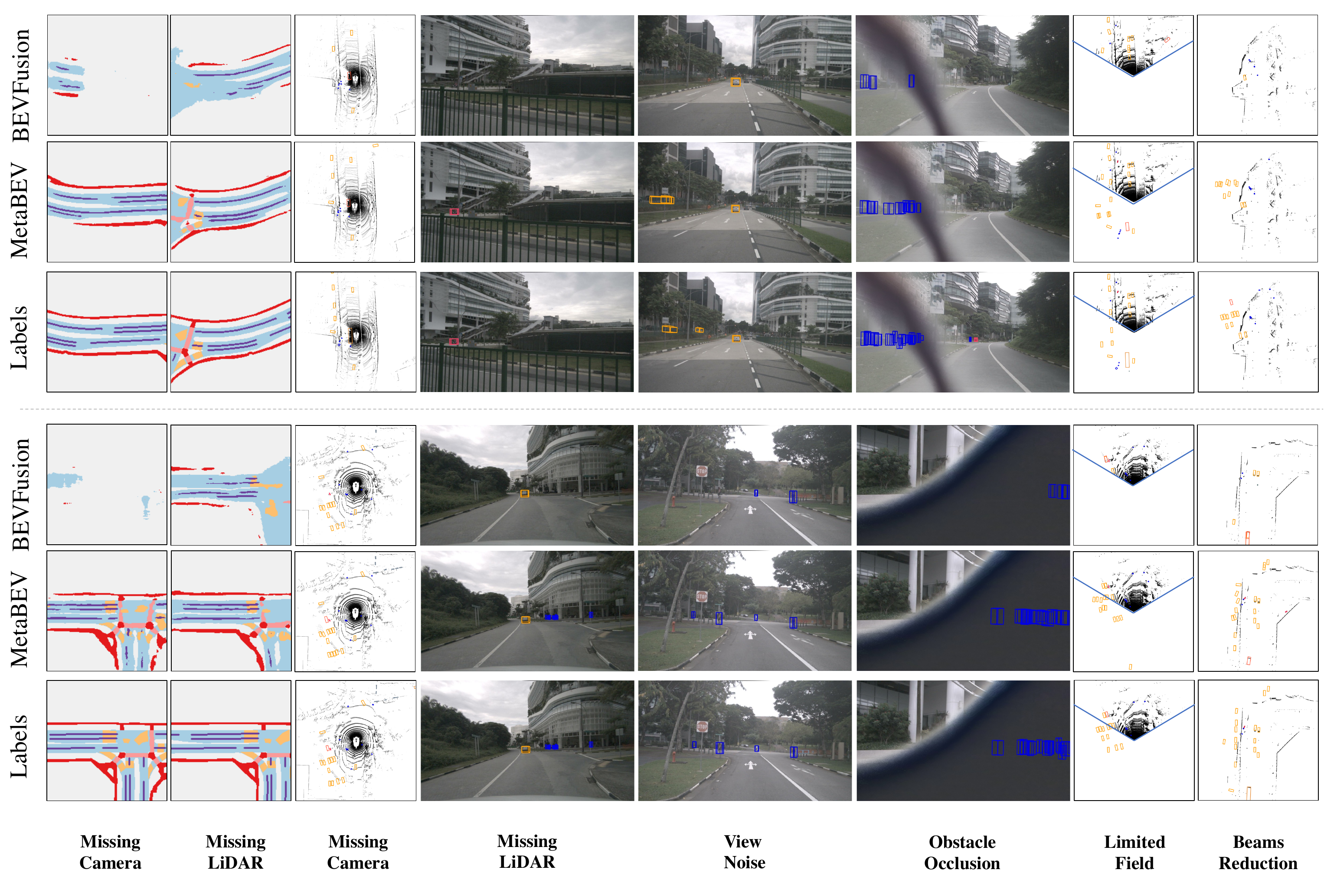}\\

\caption{\textbf{Additional 3D prediction results under various sensor failures.} 
} \label{fig:appendix_vis}
\end{center} 
\vspace{-2.5em}
\end{figure*}

As described in Sec.~\ref{sec:supp_implementation}, we formulate different degrees for each type of corruption. In the main paper, we employ the most severe (or representative) degree on each specific corruption for evaluation. To demonstrate the robustness of \model\ more comprehensively, we evaluate \model\ on 3D detection and map segmentation under various degrees of corruptions. The experimental results shown in Table~\ref{tab:appendix_corruptions} prove that \model\ outperforms BEVFusion~\cite{liu-2022-bevfusion} in zero-shot evaluation for various degrees of corruptions, with the exception of Beam Reduction, where \model\ performs comparably. Besides, the carefully designed architecture facilitates effective training of \model\ on various degree corruption patterns, making it more capable of generalization and demonstrate stronger overall performance during in-domain evaluations. Specially, when dropping 6 views, \model\ surpasses BEVFusion by 39.4\% mIoU (43.2\% \textit{v.s.} 4.1\%). It indicates \model\ is able to process arbitrary modalities for strong robustness.

\subsection{Additional Results on Map Segmentation} \label{subsec:c3-noise_mapseg}

For segmentation, there is a representative benchmark~\cite{kamann-20cvpr-cityscape_c} proposing overall sixteen corruptions to evaluate the zero-shot robustness of models, including different illumination, input noise, etc. In this appendix, to further test the robustness of \model\ , we selected 12 most commonly-encountered corruptions mentioned in~\cite{kamann-20cvpr-cityscape_c}, including cameras blur (i.e., defocus blur, glass blur, Gaussian blur), lighting conditions (i.e, brightness, contrast, saturation, JPEG compression), weather (i.e., snow, spatter, fog, frost), \etc, and generated nuScenes-C for testing the performance of \model. For comparisons, we select the representative method BEVFusion~\cite{liu-2022-bevfusion} for testing. We adopt two metrics, including the mean Intersection over Union (mIoU) for zero-shot segmentation, and the retention ratio $\rho$ between the performance on full modalities and corrupted modalities, which is calculated as follows:
\begin{equation} \label{eq:retention} 
    \rho = \frac{F_{mIoU}}{C_{mIoU}},
\end{equation} 
where $F_{mIoU}$ is the mIoU tested on the full modalities, while $C_{mIoU}$ is the one calculated on the corrupted inputs. Figure~\ref{fig:radar_mapSeg} and Table~\ref{tab:noise_on_seg} show the comparison results, which indicate that \model\ almost consistently surpasses BEVFusion~\cite{liu-2022-bevfusion} on the metrics of mIoU and retention $\rho$.

\subsection{Additional Visualizations} \label{subsec:c4-visualizations}

\begin{table*}[t]
\begin{subtable}[b]{1.0\textwidth}
\begin{center}
\resizebox{\textwidth}{!}{

    \begin{tabular}{c |c | cc  cc  cc  cc | cc cc}
    \toprule
        \multirow{3}{*}{Methods} & \multirow{3}{*}{Evaluation} & \multicolumn{2}{c}{Limited Field} & \multicolumn{2}{c}{Limited Field} & \multicolumn{2}{c}{Limited Field} & \multicolumn{2}{c}{Limited Field} & \multicolumn{2}{|c}{Obstacle Occlusion} & \multicolumn{2}{c}{Obstacle Occlusion } \\
        & & \multicolumn{2}{c}{\gain{[-180,180]}} & \multicolumn{2}{c}{\gain{[-120,120]}} & \multicolumn{2}{c}{\gain{[-90,90]}} & \multicolumn{2}{c}{\gain{[-60,60]}} & \multicolumn{2}{|c}{\gain{w/o Occlusion}} & \multicolumn{2}{c}{\gain{w Occlusion}} \\
        \cmidrule[0.2mm](r){3-4}\cmidrule[0.2mm](r){5-6}\cmidrule[0.2mm](r){7-8}\cmidrule[0.2mm](r){9-10}\cmidrule[0.2mm](r){11-12}\cmidrule[0.2mm](r){13-14}
        & & NDS & mIoU & NDS & mIoU & NDS & mIoU & NDS & mIoU & NDS & mIoU & NDS & mIoU \\
    \midrule
        BEVFusion~\cite{liu-2022-bevfusion} &  \multirow{2}{*}{zero-shot} & 71.4 & 62.3 & 52.8 & 51.9 & 49.6 & 49.6 & 41.6 & 47.2 & 71.4 & 62.3 & 68.6 & 45.7 \\     
        MetaBEV & & 71.5 & 69.3 & 57.5 & 63.4 & 52.6 & 62.3 & 47.0 & 61.1 & 71.5 & 69.3 & 70.0 & 61.1 \\   
    \midrule
        BEVFusion~\cite{liu-2022-bevfusion} &  \multirow{2}{*}{in-domain} & 65.3 & 59.6 & 48.8 & 57.0 & 47.1 & 56.3 & 42.6 & 55.5 & 70.2 & 60.5 & 69.7 & 59.3 \\   
        MetaBEV & & 71.1 & 65.8 & 61.7 & 63.1 & 58.2 & 62.4 & 54.3 & 62.1 & 70.8 & 70.6 & 70.3 & 70.2 \\   
    \bottomrule
    \end{tabular}
    }
\end{center}
\vspace{-4mm}
\caption{Experimental comparisons on \textit{Limited Field}  and \textit{Obstacle Occlusion}.}
\label{tab:appendix_corruptions_1}
\vspace{2mm}
\end{subtable}

\begin{subtable}[b]{1.0\textwidth}
\begin{center}
\resizebox{\textwidth}{!}{
    \begin{tabular}{c |c | cc  cc  cc  cc  cc cc}
    \toprule
        \multirow{3}{*}{Methods} & \multirow{3}{*}{Evaluation} & \multicolumn{2}{c}{Missing Objects} & \multicolumn{2}{c}{Missing Objects} & \multicolumn{2}{c}{Missing Objects} & \multicolumn{2}{c}{Missing Objects} & \multicolumn{2}{c}{Missing Objects} & \multicolumn{2}{c}{Missing Objects} \\
        & & \multicolumn{2}{c}{\gain{0.0 rate}} & \multicolumn{2}{c}{\gain{0.1 rate}} & \multicolumn{2}{c}{\gain{0.3 rate}} & \multicolumn{2}{c}{\gain{0.5 rate}} & \multicolumn{2}{c}{\gain{0.7 rate}} & \multicolumn{2}{c}{\gain{1.0 rate}} \\
        \cmidrule[0.2mm](r){3-4}\cmidrule[0.2mm](r){5-6}\cmidrule[0.2mm](r){7-8}\cmidrule[0.2mm](r){9-10}\cmidrule[0.2mm](r){11-12}\cmidrule[0.2mm](r){13-14}
        & & NDS & mIoU & NDS & mIoU & NDS & mIoU & NDS & mIoU & NDS & mIoU & NDS & mIoU \\
    \midrule
        BEVFusion~\cite{liu-2022-bevfusion} &  \multirow{2}{*}{zero-shot} & 71.4 & 62.3 & 70.6 & 62.3 & 68.8 & 62.2 & 67.1 & 62.2 & 65.2 & 62.1 & 62.1 & 62.1 \\   
        MetaBEV & & 71.5 & 69.3 & 70.8 & 69.2 & 69.3 & 69.2 & 67.6 & 69.2 & 65.9 & 69.2 & 62.6 & 69.2 \\   
    \midrule
        BEVFusion~\cite{liu-2022-bevfusion} &  \multirow{2}{*}{in-domain} & 70.3 & 62.0 & 69.6 & 62.0 & 69.5 & 62.0 & 67.6 & 62.0  & 67.7 & 61.9 & 65.2 & 61.9 \\   
        MetaBEV & & 70.9 & 69.9 & 70.6 & 69.9 & 70.3 & 69.9 & 69.7 & 69.9 & 69.4 & 69.9 & 68.5 & 69.9 \\   
    \bottomrule
    \end{tabular}
    }
\end{center}
\vspace{-4mm}
\caption{Experimental comparisons on \textit{Missing Objects}.}
\label{tab:appendix_corruptions_2}
\vspace{2mm}
\end{subtable}

\begin{subtable}[b]{1.0\textwidth}
\begin{center}
\resizebox{\textwidth}{!}{
    \begin{tabular}{c |c | cc  cc  cc  cc  cc}
    \toprule
        \multirow{3}{*}{Methods} & \multirow{3}{*}{Evaluation} & \multicolumn{2}{c}{Beam Reduction} & \multicolumn{2}{c}{Beam Reduction} & \multicolumn{2}{c}{Beam Reduction} & \multicolumn{2}{c}{Beam Reduction} & \multicolumn{2}{c}{Beam Reduction} \\
        & & \multicolumn{2}{c}{\gain{32 beams}} & \multicolumn{2}{c}{\gain{16 beams}} & \multicolumn{2}{c}{\gain{8 beams}} & \multicolumn{2}{c}{\gain{4 beams}}  & \multicolumn{2}{c}{\gain{1 beams}} \\
        \cmidrule[0.2mm](r){3-4}\cmidrule[0.2mm](r){5-6}\cmidrule[0.2mm](r){7-8}\cmidrule[0.2mm](r){9-10}\cmidrule[0.2mm](r){11-12}
        & & NDS & mIoU & NDS & mIoU & NDS & mIoU & NDS & mIoU  & NDS & mIoU \\
    \midrule
        BEVFusion~\cite{liu-2022-bevfusion} &  \multirow{2}{*}{zero-shot} & 71.4 & 62.3 & 64.6 & 59.3 & 61.5 & 56.7 & 58.3 & 55.3 & 33.1 & 43.8  \\   
        MetaBEV & & 71.5 & 69.3 & 65.0 & 67.1 & 61.2 & 64.7 & 57.7 & 64.3 & 32.2 & 57.9 \\   
    \midrule
        BEVFusion~\cite{liu-2022-bevfusion} &  \multirow{2}{*}{in-domain} & 70.1 & 61.1 & 66.7 & 59.7 & 65.8 & 59.2 & 64.9 & 59.0 & 51.2 & 54.9 \\   
        MetaBEV & & 71.3 & 68.2 & 68.0 & 67.0 & 67.2 & 66.5 & 66.6 & 66.3 & 54.6 & 62.0 \\   
    \bottomrule
    \end{tabular}
    }
\end{center}
\vspace{-4mm}
\caption{Experimental comparisons on \textit{Beam Reduction}.}
\label{tab:appendix_corruptions_3}
\vspace{2mm}
\end{subtable}

\begin{subtable}[b]{1.0\textwidth}
\begin{center}
\resizebox{\textwidth}{!}{
    \begin{tabular}{c |c | cc  cc  cc  cc  cc cc}
    \toprule
        \multirow{3}{*}{Methods} & \multirow{3}{*}{Evaluation} & \multicolumn{2}{c}{View Drop} & \multicolumn{2}{c}{View Drop} & \multicolumn{2}{c}{View Drop} & \multicolumn{2}{c}{View Drop} & \multicolumn{2}{c}{View Drop} & \multicolumn{2}{c}{View Drop} \\
        & & \multicolumn{2}{c}{\gain{1 drop}} & \multicolumn{2}{c}{\gain{2 drops}} & \multicolumn{2}{c}{\gain{3 drops}} & \multicolumn{2}{c}{\gain{4 drops}} & \multicolumn{2}{c}{\gain{5 drops}} & \multicolumn{2}{c}{\gain{6 drops}} \\
        \cmidrule[0.2mm](r){3-4}\cmidrule[0.2mm](r){5-6}\cmidrule[0.2mm](r){7-8}\cmidrule[0.2mm](r){9-10}\cmidrule[0.2mm](r){11-12}\cmidrule[0.2mm](r){13-14}
        & & NDS & mIoU & NDS & mIoU & NDS & mIoU & NDS & mIoU & NDS & mIoU & NDS & mIoU \\
    \midrule
        BEVFusion~\cite{liu-2022-bevfusion} &  \multirow{2}{*}{zero-shot} & 70.8 & 55.0 & 70.0 & 47.4 & 69.4 & 38.4 & 68.6 & 28.2 & 68.2 & 17.1 & 67.5 & 4.1 \\   
        MetaBEV & & 71.0 & 67.2 & 70.4 & 64.6 & 70.0 & 61.8 & 69.5 & 57.6 & 68.9 & 51.8 & 68.3 & 43.5 \\   
    \midrule
        BEVFusion~\cite{liu-2022-bevfusion} &  \multirow{2}{*}{in-domain} & 68.4 & 57.8 & 68.3 & 56.9 & 68.3 & 56.0 & 68.2 & 54.8 & 68.3 & 54.1 & 68.3 & 52.8 \\   
        MetaBEV &  & 70.2 & 68.2 & 70.0 & 68.2 & 69.8 & 68.1 & 69.6 & 68.0 & 69.4 & 68.0 & 69.3 & 67.9 \\   
    \bottomrule
    \end{tabular}
    }
\end{center}
\vspace{-4mm}
\caption{Experimental comparisons on \textit{View Drop}.}
\label{tab:appendix_corruptions_4}
\vspace{2mm}
\end{subtable}

\begin{subtable}[b]{1.0\textwidth}
\begin{center}
\resizebox{\textwidth}{!}{
    \begin{tabular}{c |c | cc  cc  cc  cc  cc cc}
    \toprule
        \multirow{3}{*}{Methods} & \multirow{3}{*}{Evaluation} & \multicolumn{2}{c}{View Noise} & \multicolumn{2}{c}{View Noise} & \multicolumn{2}{c}{View Noise} & \multicolumn{2}{c}{View Noise} & \multicolumn{2}{c}{View Noise} & \multicolumn{2}{c}{View Noise} \\
        & & \multicolumn{2}{c}{\gain{1 noise}} & \multicolumn{2}{c}{\gain{2 noise}} & \multicolumn{2}{c}{\gain{3 noise}} & \multicolumn{2}{c}{\gain{4 noise}} & \multicolumn{2}{c}{\gain{5 noise}} & \multicolumn{2}{c}{\gain{6 noise}} \\
        \cmidrule[0.2mm](r){3-4}\cmidrule[0.2mm](r){5-6}\cmidrule[0.2mm](r){7-8}\cmidrule[0.2mm](r){9-10}\cmidrule[0.2mm](r){11-12}\cmidrule[0.2mm](r){13-14}
        & & NDS & mIoU & NDS & mIoU & NDS & mIoU & NDS & mIoU & NDS & mIoU & NDS & mIoU \\
    \midrule
        BEVFusion~\cite{liu-2022-bevfusion} &  \multirow{2}{*}{zero-shot} & 70.7 & 56.9 & 69.8 & 51.4 & 69.1 & 45.5 & 68.2 & 39.1 & 67.8 & 33.1 & 66.9 & 25.8 \\   
        MetaBEV & & 70.9 & 67.0 & 70.4 & 64.4 & 70.0 & 61.4 & 69.5 & 57.2 & 68.9 & 51.8 & 68.3 & 45.1 \\   
    \midrule
        BEVFusion~\cite{liu-2022-bevfusion} &  \multirow{2}{*}{in-domain}  & 69.3 & 57.4 & 69.1 & 56.7 & 68.8 & 55.8 & 68.7 & 54.9 & 68.5 & 54.2 & 68.3 & 53.1\\   
        MetaBEV & & 70.2 & 68.0 &70.0 & 68.0 & 69.8 & 67.9 & 69.6 & 67.8 & 69.4 & 67.6 & 69.2 & 67.5  \\   
    \bottomrule
    \end{tabular}
    }
\end{center}
\vspace{-4mm}
\caption{Experimental comparisons on \textit{View Noise}.}
\label{tab:appendix_corruptions_5}
\vspace{2mm}
\end{subtable}

\caption{\textbf{Experimental comparisons on sensor corruptions with various degrees.} Texts in \gain{blue} denote the specific corruption degrees. \model\ consistently outperforms BEVFusion~\cite{liu-2022-bevfusion} on various sensor corruptions in both zero-shot and in-domain tests.}
\label{tab:appendix_corruptions}
\end{table*}

We present additional 3D prediction results in Figure~\ref{fig:appendix_vis}. It could be observed that \model\ performs robustly under various kinds of sensor failures. For example, we when the camera is totally missing, BEVFusion~\cite{liu-2022-bevfusion} can barely generate map segmentation, while \model\ is still capable of producing satisfactory results. The same phenomenon could also be found under the LiDAR-missing scenarios. Besides, for sensor corruptions, \model\ is also more robust than BEVFusion~\cite{liu-2022-bevfusion}. For instance, when \textit{View Noise} occurs as shown in the fifth column, BEVFusion tends to omit some objects (\eg, the vehicles in the first row and the pedestrian in the fourth row), while \model\ is almost able to cover all the objects in the ground truth. Finally, we provide \href{https://chongjiange.github.io/metabev.html}{more video demos} to better illustrate \model{} under all above mentioned sensor failure cases.

\onecolumn

\begin{algorithm}[H]
\small
\caption{\small Cross-Attention Modules.
}
\label{alg:attention}
\definecolor{codeblue}{rgb}{0.25,0.5,0.5}
\definecolor{codekw}{rgb}{0.85, 0.18, 0.50}
\lstset{
  backgroundcolor=\color{white},
  basicstyle=\fontsize{7.5pt}{7.5pt}\ttfamily\selectfont,
  columns=fullflexible,
  breaklines=true,
  captionpos=b,
  commentstyle=\fontsize{7.5pt}{7.5pt}\color{codeblue},
  keywordstyle=\fontsize{7.5pt}{7.5pt}\color{codekw},
  escapechar={|}, 
}
\begin{lstlisting}[language=python]
import torch
import torch.nn as nn

class CrossAttention(nn.Module):
  def __init__(self, d_model=264, n_heads=8, n_points=4, lidar_flag=False, camera_flag=False, **kwargs):
    super().__init__()

    if self.camera_flag:
        self.offsets_camera = nn.Linear(d_model, n_heads * n_points * 2)
        self.attention_weights_camera = nn.Linear(d_model, n_heads * n_points)

    if self.lidar_flag:
        self.offsets_lidar = nn.Linear(d_model, n_heads * n_points * 2)
        self.attention_weights_lidar = nn.Linear(d_model, n_heads * n_points)

    self.value_proj = nn.Linear(d_model, d_model)
    self.output_proj = nn.Linear(d_model, d_model)
    

  def forward(self, query, reference_points, x, camera_flag=False, lidar_flag=False):
    value = self.value_proj(x)

    if camera_flag and lidar_flag:
        offsets_camera = self.offsets_camera(query)
        offsets_lidar = self.offsets_lidar(query)
        offsets = torch.cat([offsets_camera, offsets_lidar])
        attention_weights_camera = self.attention_weights_camera(query)
        attention_weights_lidar = self.attention_weights_lidar(query)
        attention_weights = torch.cat([attention_weights_camera, attention_weights_lidar], dim=3)
    elif camera_flag and not lidar_flag:
        offsets = self.offsets_camera(query)
        attention_weights = self.attention_weights_camera(query)
    if lidar_flag and not camera_flag:
        offsets = self.offsets_lidar(query)
        attention_weights = self.attention_weights_lidar(query)

    attention_weights = F.softmax(attention_weights)
    sampling_locations = reference_points + offsets

    # MSDeformAttnFunction is implemented by DeformableAttention
    output = MSDeformAttnFunction.apply(value, sampling_locations.float(), attention_weights,
    output = self.output_proj(output)
    return x
\end{lstlisting}
\end{algorithm}

\begin{algorithm}[H]
\small
\caption{\small MoE.
}
\label{alg:moe}
\definecolor{codeblue}{rgb}{0.25,0.5,0.5}
\definecolor{codekw}{rgb}{0.85, 0.18, 0.50}
\lstset{
  backgroundcolor=\color{white},
  basicstyle=\fontsize{7.5pt}{7.5pt}\ttfamily\selectfont,
  columns=fullflexible,
  breaklines=true,
  captionpos=b,
  commentstyle=\fontsize{7.5pt}{7.5pt}\color{codeblue},
  keywordstyle=\fontsize{7.5pt}{7.5pt}\color{codekw},
  escapechar={|}, 
}
\begin{lstlisting}[language=python]
import torch
import torch.nn as nn

# HMoE: Hard MoE
class MultiTaskExpertMlp(nn.Module):
    def __init__(self, in_features=256, hidden_features=None, out_features=None, act_layer=nn.GELU(),
                 drop=0., det_flag=False, seg_flag=False, **kwargs):
        super().__init__()
        self.det = det_flag
        self.seg = seg_flag
        self.multi_tasks = det_flag and seg_flag
        
        if self.multi_tasks:
            self.det_expert = nn.Sequential(
                nn.Linear(in_features, hidden_features), act_layer, nn.Dropout(drop), 
                nn.Linear(hidden_features,  out_features), nn.Dropout(drop)
            )
            self.seg_expert = nn.Sequential(
                nn.Linear(in_features, hidden_features), act_layer, nn.Dropout(drop),
                nn.Linear(hidden_features, out_features), nn.Dropout(drop)
            )
            self.fuse = nn.Sequential(nn.Linear(out_features*2, out_features),)
            self.norm = nn.LayerNorm(out_features)
        elif self.seg:
            self.seg_expert = nn.Sequential(
                nn.Linear(in_features, hidden_features), act_layer, nn.Dropout(drop),
                nn.Linear(hidden_features, out_features), nn.Dropout(drop)
            )
        elif self.det:
            self.det_expert = nn.Sequential(
                nn.Linear(in_features, hidden_features), act_layer, nn.Dropout(drop),
                nn.Linear(hidden_features, out_features), nn.Dropout(drop)
            )

    def forward(self, x):
        b, n, c = x.shape
        if self.multi_tasks:
            det_fea = self.det_expert(x)
            seg_fea = self.seg_expert(x)
            fuse_fea = self.norm(self.fuse(torch.cat([det_fea, seg_fea], dim=2)))
            return fuse_fea
        elif self.det:
            x = self.det_expert(x)
            return x
        elif self.seg:
            x = self.seg_expert(x)
            return x


# RMoE: Routing MoE
from tutel import moe as tutel_moe
from tutel import net

class MultiTaskRouteMoEMlp(nn.Module):
    def __init__(self, in_features=256, hidden_features=None, out_features=None, drop=0., act_layer=nn.GELU(), **kwargs):
        super().__init__()
        self.act = act_layer
        gate_type = kwargs['gate_type']
        experts = kwargs['experts']

        # Implemented using Microsoft-tutel: https://github.com/microsoft/tutel
        self._moe_layer = tutel_moe.moe_dplayer(
            gate_type=gate_type,
            experts={'type': experts.type, 'count_per_node': experts.count_per_node,
                     'hidden_size_per_expert': hidden_features, 'activation_fn': lambda x: self.act(x),
                     'dropout_fn': lambda x: nn.Dropout(drop)(x)},
            model_dim=out_features,
            scan_expert_func=None,
            group=net.create_groups_from_world(group_count=1).data_group,   # used for DistributedDataParallel mode
            seeds=(1, dist.get_rank() + 1, 1),
            a2a_ffn_overlap_degree=experts.a2a_ffn_overlap_degree,
            parallel_type=experts.parallel_type,
            use_2dh=experts.use_2dh,
        )

    def forward(self, x, gate_index):

        return self._moe_layer(x, gate_index=gate_index)
\end{lstlisting}
\end{algorithm}

\clearpage

\twocolumn
{\small
\bibliographystyle{ieee_fullname}
\bibliography{egbib}
}

\end{document}